\documentclass{aims}
\usepackage{amsmath}
\usepackage{paralist}
\usepackage{graphics} 
\usepackage{epsfig} 
\usepackage{graphicx}  \usepackage{epstopdf}
\usepackage[colorlinks=true]{hyperref}

\usepackage{algpseudocode,algorithm,algorithmicx}
\usepackage{xcolor}

\hypersetup{urlcolor=blue, citecolor=red}

\def\currentvolume{X}
 \def\currentissue{X}
  \def\currentyear{200X}
   \def\currentmonth{XX}
    \def\ppages{X--XX}
     \def\DOI{10.3934/xx.xx.xx.xx}

\newcommand\Nx{{N_\mathrm{x}}}
\newcommand\Ny{{N_\mathrm{y}}}
\newcommand\Np{{N_\mathrm{p}}}
\newcommand\Ne{{N_\mathrm{e}}}
\newcommand\Na{{N_\mathrm{a}}}
\newcommand\Nc{{N_\mathrm{c}}}

\newcommand\bA{\mathbf{A}}
\newcommand\bB{\mathbf{B}}
\newcommand\bF{\mathbf{F}}
\newcommand\bQ{\mathbf{Q}}
\newcommand\bS{\mathbf{S}}
\newcommand\Ix{\mathbf{I}_\mathrm{x}}
\newcommand\bH{\mathbf{H}}
\newcommand\bR{\mathbf{R}}

\newcommand\bE{\mathbf{E}}

\newcommand\bzero{\mathbf{0}}
\newcommand\cf{\mathbf{f}}

\newcommand\bx{\mathbf{x}}
\newcommand\by{\mathbf{y}}
\newcommand\br{\mathbf{r}}
\newcommand\barx{\mkern 1.5mu\overline{\mkern-1.5mu \mathbf{x} \mkern-1.5mu}\mkern 1.5mu }

\newcommand\ceta{\boldsymbol \eta}
\newcommand\bphi{\boldsymbol \phi}
\newcommand\btheta{\boldsymbol \theta}

\newcommand\T{^\top}
\newcommand\tr{\mathrm{Tr}}
\newcommand\R{\mathbb{R}}

\newcommand\onehalf{\frac{1}{2}}

\title[Bayesian inference of dynamics] 
      {Bayesian inference of chaotic dynamics by merging data assimilation, machine learning and expectation-maximization}

\author[M.~Bocquet J.~Brajard A.~Carrassi and L.~Bertino]{}


 \email{marc.bocquet@enpc.fr}
 \email{julien.brajard@nersc.no}
 \email{n.a.carrassi@reading.ac.uk}
 \email{laurent.bertino@nersc.no}

\thanks{$^*$ Corresponding author: Marc Bocquet}

\begin{document}
\maketitle

\centerline{\scshape Marc Bocquet$^*$}
\medskip
{\footnotesize
 \centerline{CEREA, joint laboratory \'Ecole des Ponts ParisTech and EDF R\&D,}
 \centerline{Universit\'e Paris-Est, Champs-sur-Marne, France}
} 

\medskip

\centerline{\scshape Julien Brajard}
\medskip
{\footnotesize
  \centerline{Nansen Environmental and Remote Sensing Center, Bergen, Norway}
  \centerline{and Sorbonne University, CNRS-IRD-MNHN, LOCEAN, Paris, France} 
}

\medskip

\centerline{\scshape Alberto Carrassi}
\medskip
{\footnotesize
 \centerline{Departement of Meteorology, University of Reading and NCEO, United-Kingdom}
 \centerline{and Mathematical Institute, Utrecht University, The Netherlands}
}

\medskip

\centerline{\scshape Laurent Bertino}
\medskip
{\footnotesize
 \centerline{Nansen Environmental and Remote Sensing Center, Bergen, Norway}
}

\bigskip

 \centerline{(Communicated by the associate editor name)}

\begin{abstract}
  The reconstruction from observations of high-dimensional chaotic dynamics such as geophysical flows is hampered by
  (i) the partial and noisy observations that can realistically be obtained,
  (ii) the need to learn from long time series of data, and (iii) the unstable nature of the dynamics.
  To achieve such inference from the observations over long time series, it has been suggested to combine data assimilation and machine learning in several ways. We show how to unify these approaches from a Bayesian perspective using expectation-maximization and coordinate descents.  In doing so, the model, the state trajectory and model error statistics are estimated all together.
Implementations and approximations of these methods are discussed.
Finally,  we numerically and successfully test the approach on two relevant low-order chaotic models with distinct identifiability.
\end{abstract}


\section{Introduction}
\label{sec:introduction}

\subsection{Data assimilation}

The primary aim of data assimilation (DA), as used in the geosciences, is to accurately infer the state of a physical system from observations and theoretical knowledge of the system, such as the governing equations \cite{carrassi2018}. DA techniques are particularly useful to forecast chaotic geofluids, such as the atmosphere, the ocean, and the climate because of (i) the intrinsic dynamical instability of the flows and, (ii) the errors made in the governing equations and their numerical implementation \cite{magnusson2013}. Like in machine learning (ML), the observation sets can be huge (especially in operational weather forecast centers). However, as opposed to machine learning, the models are built on physical principles and are often computationally very costly. 

A Bayesian view on inference is universally adopted in DA \cite{asch2016}.
Typically, one looks for the unknown trajectory $\bx_{0:K}=\left\{ \bx_0, \bx_1, \ldots, \bx_K\right\}$ of the model state, i.e. the collection of all model states
$\bx_k \in \R^\Nx$, from time $t_0$ to time $t_K$. The physical system is assumed to be observed: $\by_k$ is the vector of observations at time $t_k$, while
$\by_{0:K}=\left\{ \by_0, \by_1, \ldots, \by_K\right\}$ denotes the full set of observations within the time window $[t_0, \ldots, t_K]$.
Bayes' inference then yields
\begin{equation}
\label{eq:bayes-var}
p(\bx_{0:K} |\by_{0:K}) = \frac{p(\by_{0:K} |\bx_{0:K})p(\bx_{0:K})}{p(\by_{0:K})},
\end{equation}
where $p(\bx_{0:K} |\by_{0:K})$ is the conditional probability density function (pdf), $p(\by_{0:K} |\bx_{0:K})$ is the likelihood and is usually specified, and $p(\bx_{0:K})$ is the prior pdf which depends on the dynamical model.

Because the observations are assimilated usually soon after being acquired and because the dynamics are often unstable, the inference is usually sequential in the geosciences.
In that case, the inference can be seen as a hidden Markov problem and can be written, assuming Markovian dynamics, as a sequence of an {\em analysis} and a {\em forecast} step \cite{carrassi2018}:
\begin{subequations}
  \label{eq:bayes-seq}
  \begin{align}
    \label{eq:bayes-seq-a}
    \mbox{Analysis step:} & \qquad p(\bx_k|\by_{0:k}) \propto  p(\by_{k} |\bx_k) p(\bx_k|\by_{0:k-1}) , \\
    \label{eq:bayes-seq-b}
    \mbox{Forecast step:} &  \qquad p(\bx_{k+1}|\by_{0:k}) = \int \! \mathrm{d}\bx_k \, p(\bx_{k+1}|\bx_k) p(\bx_k|\by_{0:k}) .
  \end{align}
\end{subequations}
For this inference to be scalable to high-dimensional models, approximations have been proposed, which yield a wealth of powerful DA techniques \cite{asch2016}.
For instance, the 4D-Var method seeks the maximum a posteriori of \eqref{eq:bayes-var}, while Gaussian filtering solutions such as the ensemble Kalman filter seek an approximate solution following \eqref{eq:bayes-seq}.
See \cite{kalnay2007,bocquet2013} for a comparison of their algorithmic merits.

In a challenging context with real observations, the inference bears not only on $\bx_{0:K}$, but often also on some coefficients tuning the error statistics. These are needed to accurately specify, for instance the likelihood in \eqref{eq:bayes-seq-a}, or the Chapman-Kolmogorov transition pdf in \eqref{eq:bayes-seq-b}. It is also common to infer a selection of physical parameters of an established numerical model.
But the physical model remains key in any DA inference.

\subsection{The model as a control variable}

Recently, fostered by the success of data-driven ML techniques, it has been suggested to partially or entirely learn a model of the physical system dynamics \cite{lguensat2017} from observations. There is certainly a long way before any data-driven techniques can become a substitute to a full realistic model in the geosciences. However, the subject is definitely open and fruitful. Moreover, the approach can specifically be applied to the uncertain parameterizations used in the model \cite{reichstein2019}.

Focusing on the inference of dynamics, there are critical questions that need a clarification such as the use of partial and noisy observations, the required amount of data, or the short term predictive versus asymptotic properties of the retrieved model. These questions will be addressed and discussed in this paper, in the wake of \cite{bocquet2019,brajard2020}.

Incorporating the \emph{surrogate model}, represented here by a set of static coefficients $\bA$, as a control variable in the Bayesian formalism yields \cite{abarbanel2018,bocquet2019}
\begin{equation}
\label{eq:bayes-A}
p(\bx_{0:K}, \bA |\by_{0:K}) = \frac{p(\by_{0:K} |\bA, \bx_{0:K})p(\bx_{0:K}|\bA)p(\bA)}{p(\by_{0:K})}.
\end{equation}
Consequently, the inference of the model can be seen as a generalized DA problem.
The problem is formally close to that of the weak-constraint 4D-Var \cite{tremolet2006} but where $\bA$ is also inferred \cite{bocquet2019}.

In this paper, we will generalize this approach, primarily by incorporating the statistics of model error in the inference. By definition, these statistics are not known a priori and, yet, are expected to have a strong influence on the quality of the estimate of $\bA$.

\subsection{Surrogate model representations from machine learning}

Considering solutions of \eqref{eq:bayes-A} requires to build practical and numerically efficient representations of the model.
Such representation could be defined by the flow rate of the dynamics:
\begin{equation}
\label{eq:flow-rate}
\frac{\mathrm{d}\bx}{\mathrm{d}t} = \bphi_\bA(\bx),
\end{equation}
or by the resolvent, i.e. the integration of the flow rate from one time step $t_{k-1}$ to the next one $t_k$:
\begin{equation}
\bx_k = \bF_\bA(\bx_{k-1}) .
\end{equation}
Let us mention broad categories of representations that have been proposed in the literature.
The resolvent $\bF_\bA$ could be expanded on a set of nonlinear regressors \cite{paduart2010,brunton2016} or it could be given by a neural network (NN) \cite{park1994,dueben2018, brajard2020,vlachas2019}, or an echo state network using reservoir modeling \cite{pathak2017,pathak2018a,vlachas2019}.
As opposed to the resolvent, unveiling the flow rate \eqref{eq:flow-rate} may offer an explicit representation of the model, closer to an ordinary (or partial) differential equation system.
This could be efficiently achieved through a simple expansion on monomial regressors \cite{bocquet2019} or using a NN \cite{wang1998,fablet2018,long2018}. It has been found that NN architectures for the dynamics that mimic numerical integration schemes, and which take the form of residual NNs, are particularly adequate \cite{e2017,chang2018}. Note that most of these papers only consider fully observed systems with none or little noise in the observations.

\subsection{Problem and objectives}
\label{sec:objectives}

To address partial and noisy observations, these ML representations can be plugged into an overarching DA formalism as demonstrated in \cite{bocquet2019}.
Hence both DA and ML techniques have to be considered to solve this problem. This is critical for realistic applications.

In this paper, we will generalize this DA framework, incorporating ML representations, but also accounting for model error inherent to the retrieval of an unknown model. The surrogate model, the state trajectory and model error statistics will be estimated all together, which, to the best of our knowledge, has never been achieved before.
Moreover, we will unify what has been proposed so far in the literature that aims at learning model dynamics from partial and noisy data.

The main assumptions used in this study are:
\begin{enumerate}
\item
  the governing equations of the model are autonomous, such that the set of parameters $\bA$ does not depend on time,
\item
  the physical system to be learned from is possibly chaotic such that only sequential DA schemes are viable over long training time windows. As in geophysics, the system is spatially extended and possibly high-dimensional,
\item
  the observations one learns from are often partial and noisy.
\end{enumerate}
For the sake of simplicity, other assumptions of lesser importance will be made. For instance, the observation error statistics will be assumed to be known.

Because model errors are accounted for in the inference, our surrogate model will actually be based on the flow rate
\begin{equation}
\label{eq:flow-rate-sde}
\frac{\mathrm{d}\bx}{\mathrm{d}t} = \bphi_\bA(\bx) + \ceta(t)
\end{equation}
as a generalization to \eqref{eq:flow-rate}, where $\ceta(t)$ is a $\Nx$-dimensional Wiener process.
Nonetheless, since in practice the DA system is discretized in time, model error is only injected in between two consecutive observation updates,  as is usually done in weak-constraint variational DA, which slightly differs from the continuous stochastic differential equation view offered by \eqref{eq:flow-rate-sde}.

\subsection{Outline}

In Section \ref{sec:representation}, we will first describe possible representations for the surrogate dynamics. We will discuss symmetries to be exploited in order for our methods to be scalable to high-dimensional systems.
In Section \ref{sec:bayesian-analysis}, we will define the Bayesian DA framework needed to address the inference of the dynamics, jointly or not with the model state trajectory, while model noise will be accounted for. In Section \ref{sec:optimization-var}, we will review and discuss the optimization strategies in the case where the model noise statistics are known, while
in Section \ref{sec:optimization-em} we will propose and discuss the optimization strategies in the case where they are not known. Expectation-maximization and coordinate descent will be the two key techniques for the optimization and they will help unify the approaches proposed so far. The formalism will be illustrated and numerically tested in Section \ref{sec:numerics}.

\section{Model representation}
\label{sec:representation}

Representations required for arbitrary dynamics are likely to have too many degrees of freedom to be tractable in high-dimension, i.e. typically when the flow rate of one variable is a function of all the other variables.
However, chaotic geophysical flows have properties and symmetries that could make their representation sparse
and practical, despite exhibiting complex behavior.

\subsection{Symmetries and constraints}

The main hypothesis that we will use is that the governing equations are local.
This means that the partial differential equations of the dynamics can be written in terms of point-wise algebraic and differential operators. If the dynamical system is defined in terms of ordinary differential discretized equations, like in this paper, \emph{local} means that the flow rate of a given state variable only depends on state variables defined within the stencil of the state variable.
This property is called \emph{locality}.
Specific examples will be given below.
For high-dimensional systems, this property is critical to avoid more than linear (typically quadratic) dependence  in the number of regressors, and more generally degrees of freedom in $\bA$, needed in the representation.
Locality is indirectly connected, in many realistic physical systems, to the existence of fast decreasing error correlation in space,
which led to the success of ensemble DA localization techniques in these systems.

One can further assume that the flow rate representation is translationally invariant in space.
This drastically reduces the number of required regressors in the representation.
This property will be called (statistical) \emph{homogeneity}.
This property is not as fundamental as locality since most geophysical model are inhomogeneous. However, important physical models such as homogeneous turbulence satisfy this property.
In fact, geofluids usually entail homogeneous fluid dynamics with heterogeneous forcings. Hence an appropriate hybrid representation could involve a homogeneous one for the dynamics and a heterogeneous one for the forcings.

\subsection{Sleek representation of the dynamics}

A simple representation of the surrogate dynamics is given by $\bphi_\bA(x)=\bA\br(\bx)$, where $\bA$ is a matrix of scalar coefficients of size $\Nx\times\Np$, and $\br(\bx)$ is a vector of regressors of size $\Np$. This has been discussed in detail in \cite{bocquet2019} and we only summarize the construction of the representation here.

The simplest nonlinear and non-trivial set of regressors is given by
the constant, linear and bilinear monomials in the $\left\{\bx_n\right\}_{0\le n \le \Nx-1}$ variables.
Note that through the time integration of the flow, higher-order terms are implicitly generated, along with longer distance correlations resulting in non-trivial resolvents for the dynamics.

However, the number of such regressors scales as $N_\mathrm{x}^2/2$ which is prohibitive for high-dimensional problems.
As discussed above, we can assume locality and demand that the monomials are built with variables whose locations belong to a small stencil of
typical size $L$. The matrix $\bA$ then becomes sparse and can be redefined into a dense matrix, which we still denote $\bA$ but now of size $\Nx \times \Na$.
$\Na$ is proportional to $L^{2d}$ up to a geometrical factor, where $d$ is the dimension of the underlying physical space.
The linear dependence, instead of quadratic, of the matrix size in $\Nx$ makes this representation potentially affordable.
If, in addition, homogeneity is assumed, then $\bA$ just becomes a vector of size $\Na$. This yields a minimal but efficient representation of the flow rate.

\subsection{Neural network representation of the dynamics}

An alternative is to choose a NN representation for the flow rate $\bphi_\bA$. 
We choose here to mimic the sleek representation described earlier.
To do so, we could first (i) build a dense layer DNN$_1$ from the input $\bx \in \R^\Nx$, then (ii) form the bilinear product from this layer output DNN$_1$, then (iii) concatenate this bilinear product with DNN$_1$ to form an intermediate layer, and (iv) apply to this layer a final dense layer DNN$_2$ which outputs a flow rate $\bphi_\bA(\bx)$ in $\R^\Nx$, which depends on the weights $\bA$ of the DNNs.
Not only do we generate linear and bilinear terms but constant terms are also generated with the layers' biases.
To mimic the sleek representation, the activation can be chosen linear. But other choices are possible.
A batch norm layer can also be applied first to rescale the variability in the NN.
Note however that, as explained above, the number of weights should scale like $N_\mathrm{x}^2/2$.

When accounting for locality and homogeneity, we use convolutional layers (CNNs) instead of dense layers but follow the same architecture for the whole NN. It is represented in the first row of Figure~\ref{fig:nnd}.  In this case, the number of weights is proportional to the volume of each convolution in state space and to the number of required convolutions, and has a similar scaling to the sleek homogeneous and local representation.

When accounting for locality only, without homogeneity,  we use locally connected layers instead, so that the coefficients of the convolution depend on the location. In this case, the number of weights scales like $\Nx$.

Contrary to \cite{fablet2018}, we use convolutional layers, so that, owing to locality, we can apply our techniques to high-dimensional systems. We use a simpler NN for the flow rate than in \cite{brajard2020}, which we will integrate later on, whereas \cite{brajard2020} built a NN for the resolvent.

\subsection{Time integration and resolvent}

We integrate those flow rates to build up a resolvent needed to propagate the state in between observations' times.
This integration follows the scheme proposed in \cite{bocquet2019}.
We first use an explicit numerical scheme, typically a fourth-order Runge-Kutta scheme which is based on the flow rate $\bphi_\bA$.
This yields the resolvent $\cf_\bA$ over $\delta t$. Incorporating a neural network into an integration schema like Runge-Kutta has the property of recurrent neural networks to compose several successive neural blocks with shared weights. A major difference with structures like Long-Short Term Memory (LSTM) or Gated Recurrent Unit (GRU) is that our approach does not have a memory because we rely on the assumption that the dynamics are Markovian.

\begin{figure}
\begin{center}
\includegraphics[width=\textwidth]{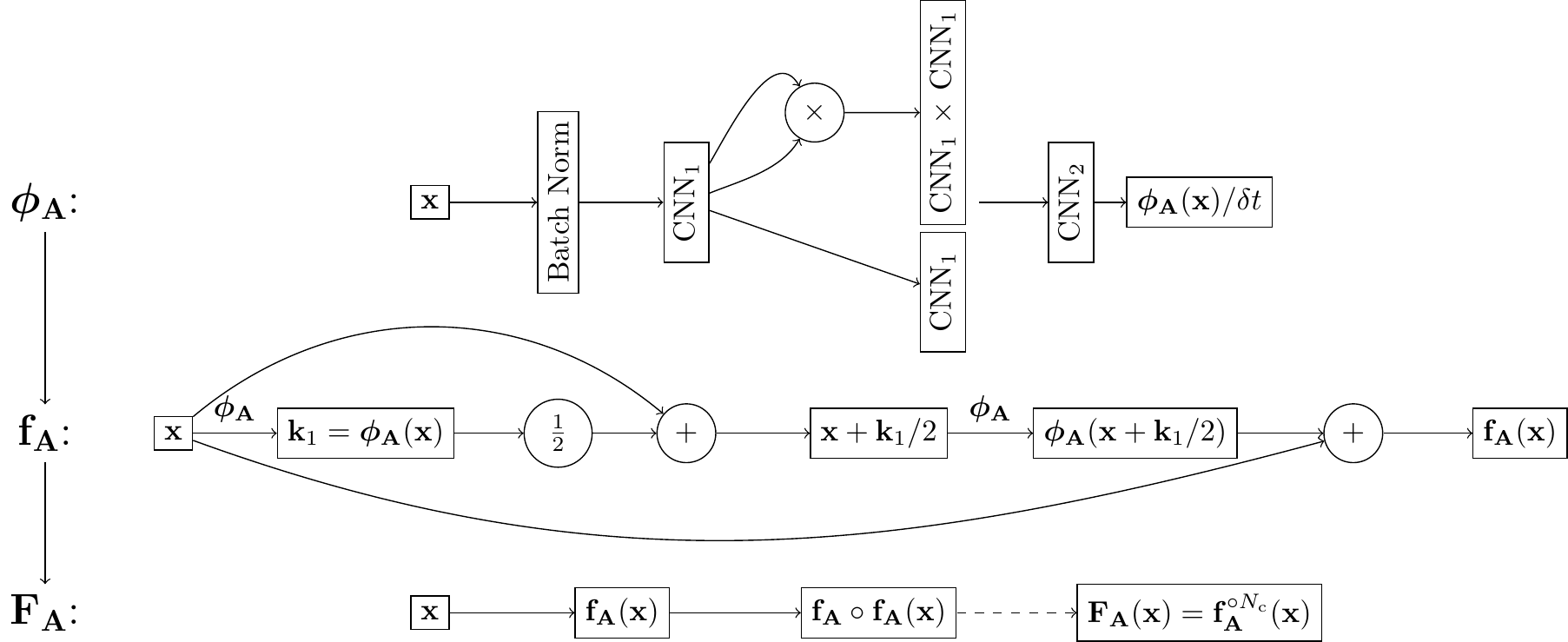}
\end{center}
\caption{\label{fig:nnd}
From top to bottom: representation of the flow rate $\bphi_\bA$ with a NN, integration of the flow rate into $\cf_\bA$ using an explicit integration scheme (here a second-order Runge Kutta scheme), and $\Nc-$fold composition up to the full resolvent $\bF_\bA$. $\delta t$ is the integration time step corresponding to the resolvent $\cf_\bA$.}
\end{figure}

We then compose this resolvent as many $\Nc$ times as required to integrate the surrogate model from $t_k$ to $t_{k+1}$, assuming $t_{k+1}-t_{k}$ is a multiple of the integration time step $\delta t$. Note that $\Nc$ could be made dependent on $k$ if the $t_k$ were not evenly spaced.
The full integration scheme, including the NN flow rate, is schematized in Figure~\ref{fig:nnd}.

For the sleek representation, its numerical integration, and its composition, we had derived in \cite{bocquet2019} the adjoint of the tangent linear model needed to formally compute the gradients of a cost function built with the resolvent.
With the NN representation, the adjoint of the resolvent needed to compute the gradients are generated by a ML tool library (typically TensorFlow or PyTorch).
Note that we are also able to efficiently build the resolvent of the sleek representation using a specific NN representation and a ML library, although this approach is not pursued here.

In the numerical illustrations of Section \ref{sec:numerics}, the NNs, and their resolvent are implemented using Keras 2.2 and TensorFlow 1.5 \cite{chollet2017}.

\section{Bayesian analysis}
\label{sec:bayesian-analysis}

At this point, we have obtained a complete representation of the resolvent of the surrogate model and its explicit dependence on the set of weights $\bA$.
Adopting a Bayesian view on the problem, we can now build the conditional pdf and associated cost function that generalized \eqref{eq:bayes-A} as introduced in Section \ref{sec:introduction}, by additionally accounting for the model error statistics.

\subsection{Prior error statistics}

Let us make assumptions on the statistics of the errors, meant to build a tractable cost function. However, other error distributions could be considered. The observation error pdf is assumed Gaussian:
\begin{equation}
\label{eq:obser}
p(\by_k|\bx_k,\bR_k) = \frac{\exp \left( -\onehalf \| \by_k - \bH_k(\bx_k)\|^2_{\bR^{-1}_k}\right)}{\sqrt{(2\pi)^{\Ny}\left| \bR_k\right|}} ,
\end{equation}
where the vector norm is generically defined by $\| \bx\|^2_\bB = \bx\T \bB \bx$.
In this paper, the observation error covariance matrices $\bR_{0:K}=\left\{\bR_0, \bR_1, \cdots, \bR_K\right\}$ are supposed to be known.
We could incorporate them into the hyperparameters to control, but this would distract us from the main focus which is model error statistics.
We also make the assumption of a Gaussian distribution for model error:
\begin{equation}
\label{eq:moder}
p(\bx_k|\bx_{k-1},\bA,\bQ_k) = \frac{\exp \left( -\onehalf \| \bx_k-\bF^{k-1}_\bA(\bx_{k-1})\|^2_{\bQ^{-1}_k}\right)}{\sqrt{(2\pi)^{\Nx}\left| \bQ_k\right|}} ,
\end{equation}
where $\bQ_{1:K}=\left\{\bQ_1, \bQ_2, \cdots, \bQ_K\right\}$ are not necessarily known.
We further assume that these Gaussian errors are white in time and that the observation and model errors are mutually independent.

As explained in \cite{bocquet2019}, there are two types of goals, one that focuses on the joint estimation of $\bA$ and $\bx_{0:K}$, and the other that focuses on the estimation of $\bA$ only.

\subsection{Joint estimation of the model, its error statistics and the state trajectory}

With the first objective, one looks for the maximum a posterior (MAP) of the joint pdf in $\bA$ and $\bx_{0:K}$ conditioned on the observations.
However, here we wish to estimate model error statistics as well.
As a consequence, we focus on the distribution of $\bA, \bQ_{1:K}, \bx_{0:K} | \mathbf{y}_{0:K}, \bR_{0:K}$.
The generalized conditional pdf to consider in this approach can be expressed using the hierarchy:
\begin{align}
  p(\bA, \bQ_{1:K}, \bx_{0:K} | &\mathbf{y}_{0:K}, \bR_{0:K}) = \nonumber\\
&  \frac{p(\mathbf{y}_{0:K}|\bx_{0:K},\bR_{0:K} )p(\bx_{0:K}|\bA, \bQ_{1:K})p(\bA, \bQ_{1:K})}{p(\mathbf{y}_{0:K}, \bR_{0:K})} ,
\end{align}
where the mutual independence of the observation and model error was used.
The first term in the numerator of the right-hand side is the likelihood of the observations and can be obtained using \eqref{eq:obser}.
The second term in the numerator is the prior on the trajectory when the model is known, which can be obtained using the Markov property of the surrogate dynamics and \eqref{eq:moder}, together with the absence of correlation of model error in time. The final term of the numerator is the joint prior of the model and the model error statistics.
Here we will not make any prior assumption on $\bA$. 
However, we will discuss this point to a limited extent in Section \ref{sec:hyperprior}, together with the use of an hyperprior for $\bQ_{1:K}$.

As a consequence, looking for the MAP of this conditional pdf, a cost function can be derived:
\begin{align}
  \label{eq:J1}
  \mathcal{J}(\bA,\bx_{0:K},\bQ_{1:K}) =& -\ln p(\bA, \bQ_{1:K}, \bx_{0:K} | \mathbf{y}_{0:K}, \bR_{0:K}) \nonumber\\
  =& \frac{1}{2}\sum_{k=0}^{K} \left\{ \left\| \mathbf{y}_k-\mathbf{H}_k(\bx_k) \right\|^2_{\mathbf{R}^{-1}_k} + \ln \left|\bR_k \right| \right\} \nonumber\\
  &  + \frac{1}{2}\sum_{k=1}^{K} \left\{ \left\| \bx_k-\mathbf{F}^{k-1}_\bA(\bx_{k-1}) \right\|^2_{\mathbf{Q}^{-1}_k}
+ \ln \left|\bQ_k \right| \right\} \nonumber\\
& - \ln p(\bx_0, \bA, \bQ_{1:K}) ,
\end{align}
up to an irrelevant constant term.
Note the resemblance of \eqref{eq:J1} with the weak-constraint 4D-Var cost function of classical DA \cite{tremolet2006}.

Very importantly, this Bayesian formulation allows for a rigorous treatment of partial and noisy observations.
The classical ML loss function that uses noiseless complete observations of the physical system can be derived from this cost function,
assuming that $\bQ_k$ is known, $\bH_k \equiv \Ix$ and letting $\bR_k$ goes to $\bzero$, so that $\mathcal{J}(\bA,\bx_{0:K},\bQ_{1:K})$ becomes in this limit
\begin{align}
  \mathcal{J}(\bA) = \frac{1}{2}\sum_{k=1}^{K} \left\| \mathbf{y}_k-\mathbf{F}^{k-1}_\bA(\mathbf{y}_{k-1}) \right\|^2_{\mathbf{Q}^{-1}_k}  - \ln p(\mathbf{y}_0,\bA,\bQ_{1:K}).
\end{align}
This view of the connections and similarities between DA and ML is similar to and extents to broader configurations than the DA and ML connections put forward in \cite{hsieh1998,abarbanel2018,bocquet2019}.

\subsection{Joint estimation of the model and the error statistics}

The second approach with a different objective is to obtain a MAP for the surrogate model, irrespective of any model state realization, i.e. we are interested in the MAP of the marginal conditional pdf
\begin{equation}
\label{eq:p2}
p(\bA, \bQ_{1:K} | \mathbf{y}_{0:K}, \bR_{0:K}) = \int \! \mathrm{d}\bx_{0:K} \, p(\bA, \bQ_{1:K}, \bx_{0:K} | \mathbf{y}_{0:K}, \bR_{0:K}) ,
\end{equation}
which is theoretically obtained by minimizing
\begin{equation}
\label{eq:J2}
\mathcal{J}(\bA, \bQ_{1:K}) = -\ln p(\bA, \bQ_{1:K} | \mathbf{y}_{0:K}, \bR_{0:K}).
\end{equation}
As pointed out in \cite{bocquet2019}, the marginal pdf \eqref{eq:p2} can be approximately related to the joint pdf through a Laplace approximation of the integral.
Here, however, we are interested in the full solution to this problem. The main optimization technique described in this paper is designed to solve this marginal problem,
and the numerical experiments will use this approach.

\section{Optimization with known model error statistics}
\label{sec:optimization-var}
We now focus on efficient numerical optimization techniques to obtain the MAP of either \eqref{eq:J1} or \eqref{eq:J2}.
In this section, we specifically focus on the joint estimation of $\bA$ and $\bx_{0:K}$,
i.e. on the cost function \eqref{eq:J1} assuming that we know $\bQ_{1:K}$, i.e. a cost function of the form ${\mathcal J}(\bA,\bx_{0:K})$ is minimized.
We unify the approaches advocated in \cite{bocquet2019,brajard2020,nguyen2019} by seeing them as
many approximate optimization techniques for the same Bayesian problem.
We discuss the pros and cons of each approach.

\subsection{Joint variational solution}
In \cite{bocquet2019}, \eqref{eq:J1} is jointly minimized on $\bA$ and on $\bx_{0:K}$, considering $\bQ_{1:K}$ to be known.
The quasi-Newton L-BFGS technique \cite{byrd1995} is chosen to implement the minimization.
The technique is very fast and efficient, provided the training window is not too long.
Indeed, the technique is well suited for nonlinear optimization problems.
However, when the training window increases, i.e. when $K$ gets larger, the approximation of the inverse Hessian formed by the L-BFGS algorithm becomes a large matrix, which heavily weights on the computational efficiency.
However, if we assume ergodicity of the model state trajectory, we can assume that distant in time pieces of the trajectory are independent. This was not exploited in \cite{bocquet2019} and this explains why it becomes impractical as $K$ becomes large.

\subsection{Coordinate descent}

More generally, because of the difference in the nature of the variables $\bA$ and $\bx_{0:K}$, we advocate
a coordinate descent \cite{wright2015}, which alternates minimization on $\bA$ and $\bx_{0:K}$. This is likely to become numerically efficient as $K$ increases.

Here, we focus on optimization strategies that use coordinate descent making explicit the differences in the $\bA$
and $\bx_{0:K}$ variables.
The key idea put forward in \cite{brajard2020}, is that the minimization on $\bA$ can be addressed using ML techniques, while the minimization on $\bx_{0:K}$ can be carried out using classical DA techniques.
Note however, that the framework is that of DA. Using ML for the minimization over $\bA$ merely means that we rely on ML libraries to solve
a variational problem because we leverage the fact that the surrogate model is a NN that can conveniently be implemented with those libraries.
Hence the use of ML is here more technical than conceptual. Let us discuss in the following subsections of the implementation of this coordinate descent optimization.

\subsection{Minimization over $\bA$}
The minimization of ${\mathcal J}(\bA, \bx_{0:K})$ over $\bA$, considering $\bx_{0:K}$ fixed, is a variational subproblem.
Because the resolvent will be implemented in Section \ref{sec:numerics} with Keras/Tensorflow, we could use stochastic gradient descent techniques,
since they are implemented in these software tools. This is what was used in \cite{brajard2020}.
We also tested this approach in the context of this paper. However, as expected with stochastic gradient descent methods,
we found the minimization to exhibit a slow convergence close to the solution.
This is fine in ML applications since accuracy is usually not an issue, or adjusted accuracy is a means to enforce regularization. Here, however, the reconstruction of the flow rate of the dynamics and its quality are sensitive to the accuracy. We found it to be more efficient to implement an L-BFGS minimizer but using the gradients generated by Keras/TensorFlow.
In particular, the convergence is faster and a better accuracy is ultimately and systematically achieved.
All the numerical experiments in Section \ref{sec:numerics} use this scheme for the minimization over $\bA$.

\subsection{Minimization over $\bx_{0:K}$}
The minimization of ${\mathcal J}(\bA, \bx_{0:K})$ over $\bx_{0:K}$, considering $\bA$, i.e. the surrogate model, to be fixed, is a classical DA problem.
Because the model is known up to some model noise, it is actually a weak-constraint data assimilation problem \cite{tremolet2006}.
It can be approximately solved using weak-constraint 4D-Var optimization techniques or using sequential schemes such as an ensemble Kalman smoother that accounts for model errors.

\subsubsection{With variational subproblems}
As explained above, using a global quasi-Newton method is prohibitive for long training window $[t_0, \ldots, t_K]$.
There are however techniques that have been proposed to parallelize the problem in time \cite{fisher2017,rao2016}.
We have actually tested numerically a variant where L-BFGS is applied in parallel over subwindows of $[t_0, \ldots, t_K]$ with possible overlapping.
This enables training over long time windows.
However, we do not wish to emphasize this approach here since it does not allow to account for uncertain model error statistics, without modifications, while ensemble techniques do.

\subsubsection{With ensemble data assimilation}
An alternative is to use an ensemble Kalman filter or smoother to estimate $\bx_{0:K}$.
In \cite{brajard2020}, an ensemble Kalman filter (EnKF, \cite{evensen2009}) is chosen. It accounts for sampling and model error through adaptive inflation \cite{bocquet2011} and additive stochastic model noise. The solution of this subproblem is the posterior mean trajectory of the EnKF.
Firstly, however, as opposed to what will be done in Section \ref{sec:optimization-em}, the posterior ensemble is later only used through its mean.
Secondly, to improve the accuracy of the trajectory estimate, one can use a smoother in place of the filter. Thirdly, to account for model error one can use a more accurate scheme than stochastic perturbations. These aspects will be discussed in Section \ref{sec:optimization-em}.

With the coordinate descent approach, we assumed that the model and observation error statistics are known. The DA techniques used to find $\bx_{0:K}$ in the coordinate descent could be supplemented by estimation techniques for $\bQ_{1:K}$ and/or $\bR_{0:K}$, which have been recently reviewed in \cite{tandeo2020}.
However, the main focus of this paper being on the marginal problem (which implicitly involves the state trajectory),
this key issue is addressed differently, as described in the following Section.

\section{Full joint optimization with expectation-maximization}
\label{sec:optimization-em}

In this section, we focus on the joint estimation of $\bA$ and $\bQ_{1:K}$, knowing the observations $\by_{0:K}$ and the statistics of their errors $\bR_{0:K}$.
Hence, we look for the MAP of \eqref{eq:p2}, i.e. a minimum of \eqref{eq:J2}.
Hereafter, we assume $\bQ_k \equiv \bQ$, i.e. that the model error statistics do not depend on time.
Estimating $\bQ_k$ at each time step would yield a severely unconstrained estimation problem.
This assumption can also be seen as a direct consequence of the hypothesis of autonomous dynamics (flawed or not), see Section \ref{sec:objectives}.
In the following, $\bQ$ will either be assumed non-parameterized (i.e. full $\bQ$) or proportional to the identity matrix $\bQ = q \Ix$. In the absence of statistical homogeneity, a diagonal $\bQ$ would also be a convenient parameterization
\cite{liu2017}, but is not tested here.
It has been suggested by \cite{ghahramani1999}, in the context where both model error statistics and a model are to be identified, that the MAP of pdfs similar to \eqref{eq:p2} could be found using the expectation-maximization (EM) technique. More recently \cite{pulido2018} have used it in a DA problem to find model and observation error statistics in the case where model error is parameterized and stochastic.

\subsection{Principle of expectation-maximization}

The EM method \cite{dempster1977} is ideally suited to find the maximum likelihood or MAP of a marginal pdf, over some parameters. The principle of the method is as follows.
We are interested in a local maximum of the conditional pdf depending on a vector $\btheta$ of parameters:
\begin{equation}
\label{eq:em-likelihood}
p(\btheta|\by) = \frac{p(\by|\btheta)p(\btheta)}{p(\by)}
= \frac{p(\btheta)}{p(\by)} \int \! {\mathrm d}\bx \, p(\by|\bx,\btheta)p(\bx|\btheta),
\end{equation}
where an expression for $p(\by|\bx, \btheta)$ and for $p(\bx|\btheta)$ is known whereas the integral being intractable, an analytic expression for $p(\by|\btheta)$ is not known.
Usually, the method focus on a likelihood $p(\by|\btheta)$ whereas here, as in \cite{liu2017}, we are interested in applying EM to the conditional pdf $p(\btheta|\by)$. As seen in \eqref{eq:em-likelihood}, the only difference is in the hyperprior $p(\btheta)$. The classical approach implicitly chooses it to be $p(\btheta)\equiv 1$. We will see in Section \ref{sec:hyperprior} that other choices are possible.

The EM algorithm, meant to find the maximum of $p(\btheta|\by)$, is iterative and alternates the following steps until convergence to a local maximum of $\btheta$.

\subsubsection{The expectation step:} Given the iterate $\btheta^{(j)}$ of the parameters
  at the iteration $j$, one computes the function ${\mathcal L}(\btheta|\btheta^{(j)}) = {\mathbb E}_{\bx|\by,\btheta^{(j)}}\left[ \ln p(\bx,\by,\btheta) \right]$, where
  ${\mathbb E}_{\bx|\by,\btheta^{(j)}}$ is the expectation operator over the distribution
  of $\bx|\by,\btheta^{(j)}$. Note that $p(\bx,\by,\btheta)=p(\by|\bx, \btheta)p(\bx|\btheta)p(\btheta)$ corresponds to the whole integrand in \eqref{eq:em-likelihood} together with the hyperprior for $\btheta$.

\subsubsection{The maximization step:} Once ${\mathcal L}(\btheta|\btheta^{(j)})$ has been computed, one can look for one of its local maxima and set it to be the new vector of parameters: $\btheta^{(j+1)} = \mathrm{argmax}_{\btheta} \, {\mathcal L}(\btheta|\btheta^{(j)})$. 

\medskip
Iterating, the EM method converges to a local maximum provided there is at least one.
An analytical expression is sometimes known for $\mathcal L$, for example when looking for the optimal mixing coefficients of a Gaussian mixture model \cite{bishop2006}.
Unfortunately, there is no known expression in our case.
An approximation meant to compute ${\mathcal L}(\btheta|\btheta^{(j)})$ estimates the expectation from a Monte Carlo sampling \cite{wei1990}.
If there are $\Ne$ such samples, an estimator for $\mathcal L$ is
\begin{equation}
{\mathcal L}(\btheta|\btheta^{(j)}) \approx \frac{1}{\Ne}\sum_{i=1}^\Ne \ln p(\bx_i^{(j)},\by,\btheta),
\end{equation}
where $\bx_i^{(j)}$ is a sample of $\bx | \by, \btheta^{(j)}$.

Alternatively, one can look for a local minimum of $-{\mathcal L}(\btheta|\btheta^{(j)})$.  This is the convention adopted in the following and we redefine $-{\mathcal L}$ as ${\mathcal L}$, and deal with minima instead of maxima.

\subsection{Full scheme}
Hence, getting the MAP of \eqref{eq:p2} can in principle be numerically solved using an EM-like algorithm.
In our problem, $\btheta$ is the set of parameters $(\bA, \bQ)$, while $\by$ is $\by_{0:K}$ and the marginal integration is over all possible trajectories $\bx_{0:K}$.

We choose for the initial $\bA^{(0)}$ a small set of random coefficients. We specifically want these to be small so that $\bA^{(0)}$ lies not too far from the domain of stability of the generated dynamics; see Section 3.2 of \cite{bocquet2019} for a discussion.
We choose $\bQ^{(0)}$ to be the matrix $q^{(0)}\Ix$ where $q^{(0)}$ will be chosen large enough so as not to underestimate model error and avoid a premature divergence in the ensemble DA step. For sequential DA schemes, an initial $\bx^{(0)}$ also has to be chosen. The final iterates should not depend on these initial values except for picking the local minimum towards which the algorithm converges.

\medskip
\subsubsection{The expectation step:} We assume $\bA, \bQ$ to be given by their estimates at iteration $j$, i.e. $\bA^{(j)}, \bQ^{(j)}$.
    The expectation step requires an ensemble DA technique, which later enables the use of Monte Carlo EM, the ensemble providing the samples.
    Specifically, we will rely in the numerical experiments of Section \ref{sec:numerics} on the classical ensemble Kalman smoother (EnKS, \cite{evensen2009}) based on the ensemble transform formulation \cite{bishop2001,hunt2007}. The smoother has two steps: the first one is a filtering pass; the second one is the smoother pass which traces back $L$ time steps in the past, where $L$ is known as the lag of the smoother.
    The model used in the EnKS is by construction the surrogate model specified by $\bA^{(j)}$.
    
    In order not to be annoyed by difficult technical details of the implementation which are not key to the method presented here, we will assume that the size of the ensemble $\Ne$ is large enough so that (i) localization is not necessary (ii) model error is well represented by an ensemble with a large span. For instance if $\Ne-1$ is chosen to be the dimension of the surrogate model state vector, on can expect on the basis of the experiments presented in \cite{raanes2015} that the posterior error is minimal.
    In realistic, high-dimensional problems where only a small ensemble can be afforded, localization would have to be used and would be efficient as demonstrated in \cite{sakov2018}.
    Model error is accounted for in the filtering pass of the EnKS using the deterministic SQRT-CORE scheme \cite{raanes2015} using $\bQ^{(j)}$. With a large ensemble size, we expect it to be more accurate than the addition of stochastic noise as in \cite{brajard2020}.
    The output of this step is the smoothed ensemble trajectory $\bE^{(j)}_{0:K} \in \R^{K\times\Nx\times\Ne}$ as computed by the EnKS.
    
\medskip
\subsubsection{The maximization step:} At iteration $j$, we assume the ensemble trajectory
\begin{equation}
\bE^{(j)}_{0:K} = \left\{ \bx^{(j)}_{i,k} \right\}_{1 \le i \le \Ne, 0\le k \le K}
\end{equation}
to be given by the expectation/DA step. Following the Monte Carlo EM algorithm, we would like to minimize
\begin{align}
\label{eq:full-em}
  {\mathcal L}^{(j)}(\bA,\bQ) =& - \frac{1}{\Ne}\sum_{i=1}^\Ne \ln p(\bE_i^{(j)}, \by_{0:K}, \bA, \bQ, \bR_{0:K}) \nonumber\\
   = & \frac{1}{2\Ne}\sum_{i=1}^\Ne\sum_{k=1}^{K} \left\{ \left\| \bx^{(j)}_{k,i}-\mathbf{F}^{k-1}_\bA(\bx^{(j)}_{k-1,i}) \right\|^2_{\mathbf{Q}^{-1}}
  + \ln \left|\bQ \right| \right\} \nonumber\\
  & - \ln p(\bx^{(j)}_{0,i}, \bA, \bQ) + \cdots ,
\end{align}
up to an irrelevant constant term.

\subsubsection{Joint optimization of $\bA$ and $\bQ$}
\label{sec:joint-optimization-A-Q}

A straightforward solution of the maximization step consists in optimizing ${\mathcal L}(\bA,\bQ)$ on both $\bA$ and $\bQ$ using L-BFGS of \cite{byrd1995} and the gradients generated by an ML library such as Keras/Tensorflow. Note that this requires to store the full ensemble of trajectories over the training window.

The full scheme algorithm is summarized in Algorithm~\ref{alg:algo-em-fs}.
The criteria for convergence can be rather complex, having to monitor both $\bA$ and $\bQ$.
For the sake of simplicity, a  fixed number of iterations $N_\mathrm{iter}$ will be imposed in Section \ref{sec:numerics}.

\begin{algorithm}
  \caption{Full scheme Monte Carlo EM algorithm for estimating $\bA$ and $\bQ$ \label{alg:algo-em-fs}}
  \begin{algorithmic}[1]
    \State $j=0$
    \State Set $\bA^{(0)}$ and $\bQ^{(0)}$
    \While{ ($\bA^{(j)}$ or $\bQ^{(j)}$ are not converged) and $j<N_\mathrm{iter}$}
      \State Expectation/DA step: EnKS with SQRT-CORE using $\bA^{(j)},\bQ^{(j)}$ \par
      Output: $\bE^{(j)}_{0:K}$
      \State 
      Maximization/ML step: Minimize ${\mathcal L}^{(j)}(\bA,\bQ)$ knowing $\bE^{(j)}_{0:K}$ \par
      Output: $\bA^{(j+1)}$ and $\bQ^{(j+1)}$
      \State $j \longleftarrow j+1$ 
      \EndWhile
  \end{algorithmic}
\end{algorithm}

We tested this option of jointly optimizing on $\bA$ and $\bQ$ with success.
However, the minimization of ${\mathcal L}^{(j)}(\bA,\bQ)$ over $\bQ$ alone has an analytic expression which can efficiently be leveraged upon.
To do so, one could use a coordinate descent on $\bA$ and $\bQ$.
At iteration $j$ of the EM algorithm, one first minimizes ${\mathcal L}(\bA,\bQ^{(j)})$ using $\bA^{(j)}$ as starting point of the minimization.
Note that this cost function is essentially the classical least squares loss:
\begin{equation}
\label{eq:least-squares}
   {\mathcal L}^{(j)}(\bA,\bQ^{(j)}) = \frac{1}{2\Ne}\sum_{i=1}^\Ne\sum_{k=1}^{K}  \left\| \bx^{(j)}_{k,i}-\mathbf{F}^{k-1}_\bA(\bx^{(j)}_{k-1,i}) \right\|^2_{{\mathbf{Q}^{(j)}}^{-1}},
   \end{equation}
where the prior on $\bA$ is neglected.
The local minimum which is obtained is noted $\bA^{(j,1)}$. Then one can solve for the exact minimum of ${\mathcal L}(\bA^{(j,1)},\bQ)$ over the non-parameterized $\bQ$:
\begin{equation}
\label{eq:Q-em}
\bQ^{(j,1)} = \frac{1}{K\Ne}\sum_{i=1}^\Ne\sum_{k=1}^K \left( \bx^{(j)}_{k,i}-\mathbf{F}^{k-1}_{\bA^{(j,1)}}(\bx^{(j)}_{k-1,i}) \right)
\left( \bx^{(j)}_{k,i}-\mathbf{F}^{k-1}_{\bA^{(j,1)}}(\bx^{(j)}_{k-1,i})\right)\T .
\end{equation}
Similar expressions can be obtained in the case where $\bQ$ is diagonal or of the specific form $q\Ix$ (see e.g., \cite{liu2017}).
With this new $\bQ^{(j,1)}$, on can go back to the minimization on $\bA$ with the initial guess being $\bA^{(j,1)}$ and iterate the coordinate descent until convergence.
This ultimately yields $\bA^{(j+1)}$ and $\bQ^{(j+1)}$.

However, we do not expect the number of iterations of this coordinate descent to be large. Indeed, ${\mathcal L}(\bA,\bQ^{(j)})$ as seen in \eqref{eq:least-squares} should not be very sensitive to $\bQ^{(j)}$ once $\bQ^{(j,\star)}$ is close to convergence because it only plays the role of a Mahalanobis metric in the inner product.
To support this argument, note that in the case where $\bQ = q\Ix$, the solution of the minimization of ${\mathcal L}(\bA,\bQ^{(j)})$ does not actually depends on $\bQ$ at all, since it only represents a scaling factor of the loss function.
In this case, at iteration $j$ of the EM algorithm, the coordinate descent simplifies to a single minimization of ${\mathcal L}(\bA,\bQ^{(j)})$ on $\bA$ which yields $\bA^{(j+1)}$, followed by
\eqref{eq:Q-em} which yields $\bQ^{(j+1)}$.

\subsection{Approximate scheme}
\label{sec:approximate-scheme}

We now present an approximation of the maximization step, which avoids to store the full ensemble of smoothed trajectories over the training window (as required by \eqref{eq:full-em} or \eqref{eq:least-squares}).

At iteration $j$ of the EM algorithm, we first implement the EnKS with SQRT-CORE using $\bA^{(j)}$ and $\bQ^{(j)}$.
Because we want to avoid computing $\bQ^{(j+1)}$ at the maximization step through
\begin{equation}
\label{eq:Q-em-bis}
\bQ^{(j+1)} = \frac{1}{K\Ne}\sum_{i=1}^\Ne\sum_{k=1}^K \left( \bx^{(j)}_{k,i}-\mathbf{F}^{k-1}_{\bA^{(j+1)}}(\bx^{(j)}_{k-1,i}) \right)
\left( \bx^{(j)}_{k,i}-\mathbf{F}^{k-1}_{\bA^{(j+1)}}(\bx^{(j)}_{k-1,i})\right)\T,
\end{equation}
which requires the full ensemble, we rather compute it online in the expectation step by accumulating terms of the sum as the EnKS unfolds.
Moreover, instead of extracting the full ensemble from the EnKS, we only store and use the smoothed ensemble trajectory mean $\barx^{(j)}_{0:K}$.

Still at iteration $j$ of the EM algorithm, focusing on the maximization step, we carry out the minimization of
\begin{align}
\label{eq:A-em}
  {\mathcal L}^{(j)}(\bA,\bQ^{(j+1)}) =& -\ln p(\barx^{(j)}_{0:K}, \by_{0:K}, \bA, \bQ^{(j+1)}, \bR_{0:K}) \nonumber\\
  =  & \frac{1}{2}\sum_{k=1}^{K} \left\{ \left\| \barx^{(j)}_k-\mathbf{F}^{k-1}_\bA(\barx^{(j)}_{k-1}) \right\|^2_{\mathbf{Q^{(j)}}^{-1}}
  + \ln \left|\bQ^{(j+1)} \right| \right\} \nonumber\\
  & - \ln p(\barx^{(j)}_0, \bA, \bQ^{(j+1)}) + \cdots ,
\end{align}
with respect to $\bA$, up to terms not depending on $\bA$.
The result is a local minimum: $\bA^{(j+1)} = \mathrm{argmin}_\bA \, {\mathcal L}^{(j)}(\bA,\bQ^{(j+1)})$.
We cannot iterate the minimization on $\bA$ and $\bQ$ as the $\bQ^{(j+1)}$ has been computed once and for all in the expectation step,
which is a significant approximation of the coordinate descent. However, as argued in Section \ref{sec:joint-optimization-A-Q}, this approximation should be mild since, if $\bQ=q\Ix$, the coordinate descent would not require iterating.

The approximate algorithm is summarized in Algorithm~\ref{alg:algo-em-ap}.
\begin{algorithm}
  \caption{Approximate Monte Carlo EM algorithm for estimating $\bA$ and $\bQ$ \label{alg:algo-em-ap}}
  \begin{algorithmic}[1]
    \State $j=0$
    \State Set $\bA^{(0)}$ and $\bQ^{(0)}$
    \While{ ($\bA^{(j)}$ or $\bQ^{(j)}$ are not converged) and $j<N_\mathrm{iter}$}
      \State Expectation/DA step: EnKS with SQRT-CORE using $\bA^{(j)}, \bQ^{(j)}$ \par
      Output: $\barx^{(j)}_{0:K}$ and $\bQ^{(j+1)}$ computed online
      \State 
      Maximization/ML step: Minimize ${\mathcal L}^{(j)}(\bA,\bQ^{(j+1)})$ knowing $\barx^{(j)}_{0:K}$ \par
      Output: $\bA^{(j+1)}$
      \State $j \longleftarrow j+1$ 
      \EndWhile
  \end{algorithmic}
\end{algorithm}
 
The authors of \cite{nguyen2019} have nicely suggested to use the EM algorithm to learn a model from data following \cite{ghahramani1999}, in particular using an EnKS for the expectation step. However, in their algorithms and applications, they (i) do not estimate the model error noise, (ii) do not make use  of the ensemble output of the EnKS in the maximization step, which is key to EM. As a result, even though their algorithms have the structure of EM, their algorithms are merely coordinate descents applied to ${\mathcal J}(\bA, \bx_{0:K})$, which amounts to the algorithm proposed in \cite{brajard2020}.

\subsection{Hyperpriors}
\label{sec:hyperprior}

\subsubsection{Hyperprior on $\bQ$}

An hyperprior for $\bQ$ is allowed by the general formalism \eqref{eq:em-likelihood}. We expect such an hyperprior to be impactful only in the case of short training windows; it should be overridden by increasing data.
If $\bQ$ is parameterized as $q\Ix$, we can use a scalar Jeffreys hyperprior on the variance, which is
$p(q) = q^{-1}$. Focusing only on the occurrences of $q$, the main cost function is
\begin{equation}
{\mathcal L}(\bA,\bQ) = -\ln p(q) +\frac{Ks}{2q} + \frac{K\Nx}{2}\ln(q) + \ldots
\end{equation}
where
\begin{equation}
s =  \frac{1}{K\Ne}\sum_{i=1}^\Ne\sum_{k=1}^{K} \left\| \bx_{k,i}-\mathbf{F}^{k-1}_\bA(\bx_{k-1,i})\right\|^2 .
\end{equation}
Minimizing on $q$, this yields for the maximization step:
\begin{equation}
q = \frac{K}{K\Nx+2}s .
\end{equation}
Clearly, the influence of the hyperprior on the estimate of $q$ ($+2$ in the denominator) is minor, especially for large system size or training window.

If $\bQ$ is full, we can use a matrix Jeffreys hyperprior $p(\bQ) = \left| \bQ \right|^{-\frac{\Nx+1}{2}}$.
Focusing only on the occurrences of $\bQ$, the main cost function is
\begin{equation}
{\mathcal L}(\bA,\bQ) = -\ln p(\bQ) + \frac{K}{2} \tr \left(\bS\bQ^{-1}\right) + \frac{K}{2}\ln(\left| \bQ \right|) + \ldots
\end{equation}
where
\begin{equation}
\bS =  \frac{1}{K\Ne}\sum_{i=1}^\Ne\sum_{k=1}^{K} \left( \bx_{k,i}-\mathbf{F}^{k-1}_\bA(\bx_{k-1,i})\right)
 \left( \bx_{k,i}-\mathbf{F}^{k-1}_\bA(\bx_{k-1,i})\right)\T .
\end{equation}
Minimizing on $\bQ$, this yields for the maximization step:
\begin{equation}
\bQ = \frac{K}{K+\Nx+1}\bS .
\end{equation}
The influence of the hyperprior on the estimate of $\bQ$ can be stronger in this case, in particular when the training window length is smaller or of the order of the state space dimension.

\subsubsection{Hyperprior on $\bA$}

The design of the hyperprior for $\bA$ is primarily driven by physical modeling and numerical stability.
Its role has been discussed to some extent in Section 3.1 of \cite{bocquet2019}.
Practically, such an hyperprior could be implemented by adding a regularization term (typically L1 or L2 norm) on the coefficients of $\bA$, corresponding to specific distribution assumptions for $\bA$. We avoid such regularization here by mostly considering very long training windows, and because $\bA$ is rather well constrained by locality and/or homogeneity. However, with higher dimensional physical models, larger $\bA$, deeper NN representations, and shorter training windows by comparison, methods used in machine learning and deep learning to regularize and avoid overfitting could be used, for instance dropouts and stochastic optimization techniques (e.g., Section 1.4 of \cite{lecun2012}).

\section{Numerical illustrations}
\label{sec:numerics}

In this section, we test the algorithms introduced in Section \ref{sec:optimization-em}.
The algorithms reinterpreted as coordinate descent in Section \ref{sec:optimization-em},
and for which model error statistics were prescribed have been tested in \cite{bocquet2019,brajard2020}.

\subsection{Two low-order reference models}

The new methods are tested on two low-order models: the Lorenz-96 model (L96, \cite{lorenz1998}) and the two-scale Lorenz model, sometimes known as Lorenz-05III (L05III, \cite{lorenz2005}). Both are spatially distributed one-dimensional models defined over circles.
In the following experiments, they are called the \emph{reference} models.

\subsubsection{Lorenz-96}

The L96 model is defined by a set of ODEs over a periodic domain with variables indexed by $n=0,\ldots,N_x-1$:
\begin{equation}
  \frac{{\mathrm{d}}x_n}{{\mathrm{d}}t} = (x_{n+1}-x_{n-2})x_{n-1}-x_n+F ,
\end{equation}
where $N_x=40$, $x_{N_x}=x_0$, $x_{-1} = x_{N_x-1}$, $x_{-2}=x_{N_x-2}$, and $F=8$.
This model is an idealized representation of a one-dimensional latitude band of the Earth atmosphere. Its Lyapunov time (inverse of the largest Lyapunov exponent) is $0.60$. The standard deviation of any model state variable is $3.62$. 
The truth run of the L96 model is integrated using the fourth-order Runge Kutta (RK4) scheme and with the time step $\delta t_\mathrm{r}=0.05$.

A classical DA configuration consists in the observation of every variable of the model every $\Delta t_\mathrm{r} = \delta t_\mathrm{r} = 0.05$.
The observations are perturbed with random white in time normal noise of covariance matrix $\bR = \sigma^2_y \Ix$.
In this classical configuration, we have $\sigma_y=1$.

\subsubsection{Lorenz-05III}

The two-scale Lorenz model L05III is given by the following two-scale set of ODEs:
  \begin{subequations}
    \label{eq:l05III}
    \begin{align}
      \frac{\mathrm{d}x_n}{\mathrm{d}t} &= \psi^+_n(\mathbf{x})+F-h\frac{c}{b}\sum_{m=0}^{9} u_{m+10n}, \\
      \frac{\mathrm{d}u_m}{\mathrm{d}t} &= \frac{c}{b} \psi^-_m(b\mathbf{u}) + h\frac{c}{b} x_{m/10}, \\
      \psi_n^\pm (\mathbf{x}) &= x_{n\mp 1}(x_{\pm1}-x_{n\mp 2})-u_n,
    \end{align}
  \end{subequations}
for $n=0,\ldots,N_x-1$ with $N_x=36$, and $m=0,\ldots,N_u-1$ with $N_u=360$. The indices are defined periodically over their domain; $m/10$ stands for the integer division of $m$ by $10$. The other parameters: $c=10$ for the time-scale ratio, $b=10$ for the space-scale ratio, $h=1$ for the coupling, and $F=10$ for the forcing, are set to their original values. When uncoupled ($h=0$), the Lyapunov time of the slow variables $\mathbf{x}$ sector of the model \eqref{eq:l05III} is $0.72$, which is the key time scale when focusing on the slow variables \cite{carlu2019}. The standard deviation of any of the slow variables is $3.54$. 
  
The vector $\mathbf{u}$ represents unresolved scales and hence model error when only considering the slow variables $\mathbf{x}$. It is integrated with an RK4 scheme and the time step $\delta t_\mathrm{r}=0.005$ since it is stiffer than the L96 model.

A classical DA configuration is defined by the observation of every slow variable of the model, with a time step of $\Delta t_\mathrm{r}=10 \, \delta t_\mathrm{r} = 0.05$.
The observations are perturbed with random white in time normal noise of covariance matrix $\bR = \sigma^2_y \Ix$.
In this classical configuration, we have $\sigma_y=1$.

The goal is to learn surrogate dynamics from observing these models.
Note that in the L05III case, only the slow variables are observed and only them are represented in the surrogate model and forecasted.
Given that only its slow scale is observed and reconstructed, the L05III reference model is not identifiable in that it does not belong to the space generated by all possible surrogate models. It is vain to attempt to exactly match the ODEs description of the reference models. Although less obvious, the L96 model may neither be identifiable by the NN representation. Indeed, we cannot obtain a perfect numerical reconstruction of the dynamics
to numerical precision through the NN representation, as opposed to when using the sleek representation \cite{bocquet2019}.

\subsection{Surrogate model}

For both reference models, we will use a surrogate model with a NN architecture based on the RK4 scheme. Its architecture is the same as the one depicted in Figure \ref{fig:nnd}, except for RK2 replaced by RK4. We choose $\Nc=1$, for the sake of simplicity and because the accuracy improvement offered by $\Nc > 1$, which has been tested, is often small.

The CNN$_1$ layer is a convolutional layer with a width of $5$ sites, and comprises $6$ filters (or \emph{channels}). We have checked that a larger width or more filters does not lead to noticeable improvements.
This is consistent with the theory developed in \cite{bocquet2019} which emphasizes that only a few key parameters are required (ultimately those of the ODEs of the true model).   
For CNN$_1$, we use linear activation functions. We have also tested relu and $\tanh$ activation functions, which turned out to have no significant impact or can degrade the accuracy of the surrogate model.
All occurrences of the CNN$_1$ layer within the NN share the same weights.

The CNN$_2$ layer is a point-wise convolutional layer with one filter that outputs a state vector. It always has linear activation functions.
Note that in the following experiments, batch normalization is not used.

This minimal NN (i.e. assuming locality and homogeneity) only has $49$ weights (for both reference models).
With locally connected convolutional layers, hence assuming locality but breaking translational invariance, it has $1960$ and $1764$ weights for the L96 and L05III experiments, respectively.

\subsection{Evaluation metrics}

Once the surrogate models are learned from the observations, their quality must be evaluated considering both the short-term and asymptotic properties of the dynamics. 

Our evaluation metrics for the numerical experiments will be:
\begin{itemize}
\item
Forecast skill (FS): the normalized root mean square error (NRMSE) is the root mean square difference between a forecast of the reference model and a forecast of the surrogate model
starting from the same initial condition (up to machine precision), and normalized by the standard deviation of the reference model variables. This score depends on the lead time.
The FS is averaged over $5\times 10^3$ model runs with as many independent initial conditions (not taken from the training window) to obtain a satisfactory convergence. Above a NRMSE of $1$, the forecast of the surrogate model has become useless.
\item
  Lyapunov spectrum (LS): It measures the asymptotic properties of the model and is computed numerically for both the reference and surrogate model. The spectrum is computed over $10^4$ time steps, using either finite differences
  or a Jacobian indirectly obtained from Keras/TensorFlow.
\item
Power spectrum density (PSD): It measures the energy content of the dynamics in spectral space, as a function of the wave frequency. The PSD is computed as an average on all grid points using the Welch method \cite{welch1967}.
\end{itemize}
Hovm\"oller diagrams can also be exploited to visualize differences of trajectories of the reference and the surrogate models. Some of them have been displayed in \cite{bocquet2019,brajard2020}.

Several scalar indicators will also be used.
The average validation prediction time when the NRMSE first reaches $0.5$ will be denoted $\pi_\onehalf$ (in units of Lyapunov time).
The average diagnosed residual model error will be measured by
\begin{equation}
\sigma_q = \sqrt{\frac{1}{\Nx}\mathrm{tr}(\bQ^\star)} ,
\end{equation}
where $\bQ^\star$ is the model error covariance matrix estimated in the training phase.
The average first Lyapunov exponent will be noted $\lambda_1$.

\subsection{First results}

\subsubsection{Preliminary screening of hyperparameters}
\label{sec:prelim}
We have carried out a series of experiments with the parameters described above, but taking
the EnKS lag in the set $L = \left\{0, 2, 4, 6, 8, 10 ,12\right\}$ and the training window length in the set
$K=\left\{50, 100, 200, 400, 800, 1600, 3200, 6400\right\}$. From these experiments, we picked the \emph{nominal} configuration $L=4, K=5000$ as a configuration with satisfactory forecast skills, i.e. which maximizes $\pi_\onehalf$.
Yet, such a criterion does not guarantee that the asymptotic properties indicators are optimal as well.

The fact that $L=4$, a rather small value, is close to optimal may be due to the balance between (i) a large $L$ which in principle offers
samples that are more representative of the conditional pdf, and (ii) a degradation of the smoother due to the presence of model noise as $L$ gets larger and (iii) the fact we used the classical smoother and not iterative, more optimal ones.

Some of the indicators may display variability when changing the seed of the random generator used to compute the reference model trajectory and to perturb the observations.
This point will be investigated in the following for each of the three indicators by running $10$ experiments with a new trajectory of the reference model each, and hence new observations, due to the choice of a new seed.

As mentioned before, the stopping criterion of the EM algorithm will simply be the number of iterations chosen to be $N_\mathrm{iter}=25$.
In all the numerical experiments, we observe that the sequence $\bQ^{(j)}$ usually converges in less than $10$ iterations, but that the model weights sequence $\bA^{(j)}$ keeps improving a little bit more beyond this point; hence the safe choice $N_\mathrm{iter}=25$. 

We have first compared the results of configurations using either $\bQ = q\Ix$ or the full $\bQ$. We found a slight improvement due to the full $\bQ$ representation. Since the overhead computational cost due to the use of $\bQ$ is mild for our experiments, we chose it for all the following reported experiments. Note that the MAP $\bQ$ that we obtain in the subsequent experiments are systematically close to diagonal, for both models, which supports the little difference noticed between $\bQ = q\Ix$ and the full $\bQ$.

\subsubsection{Results for the L96 reference model}

\begin{figure}
\begin{tabular}{ccc}
  \includegraphics[width=0.5\textwidth]{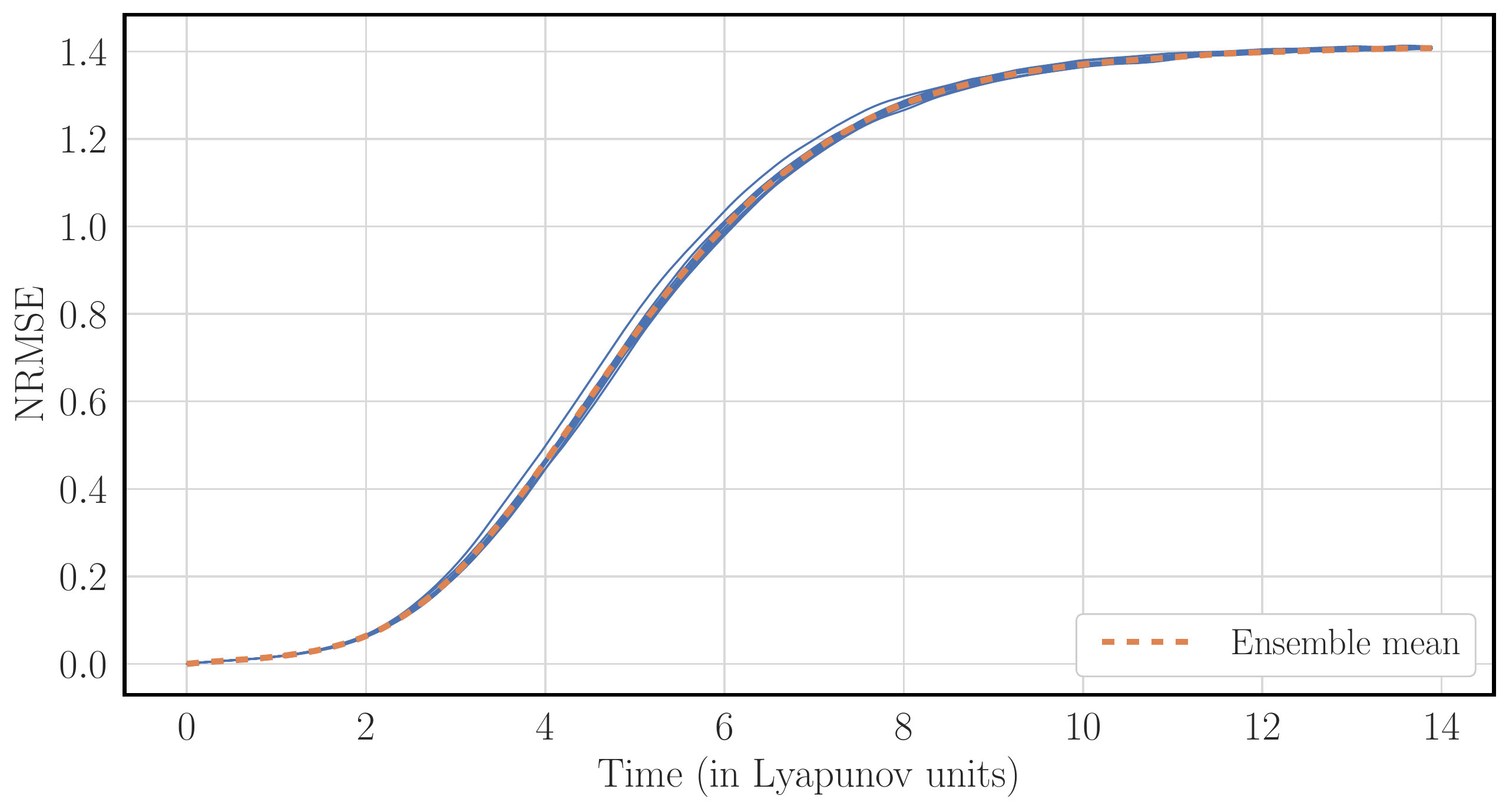} & \includegraphics[width=0.5\textwidth]{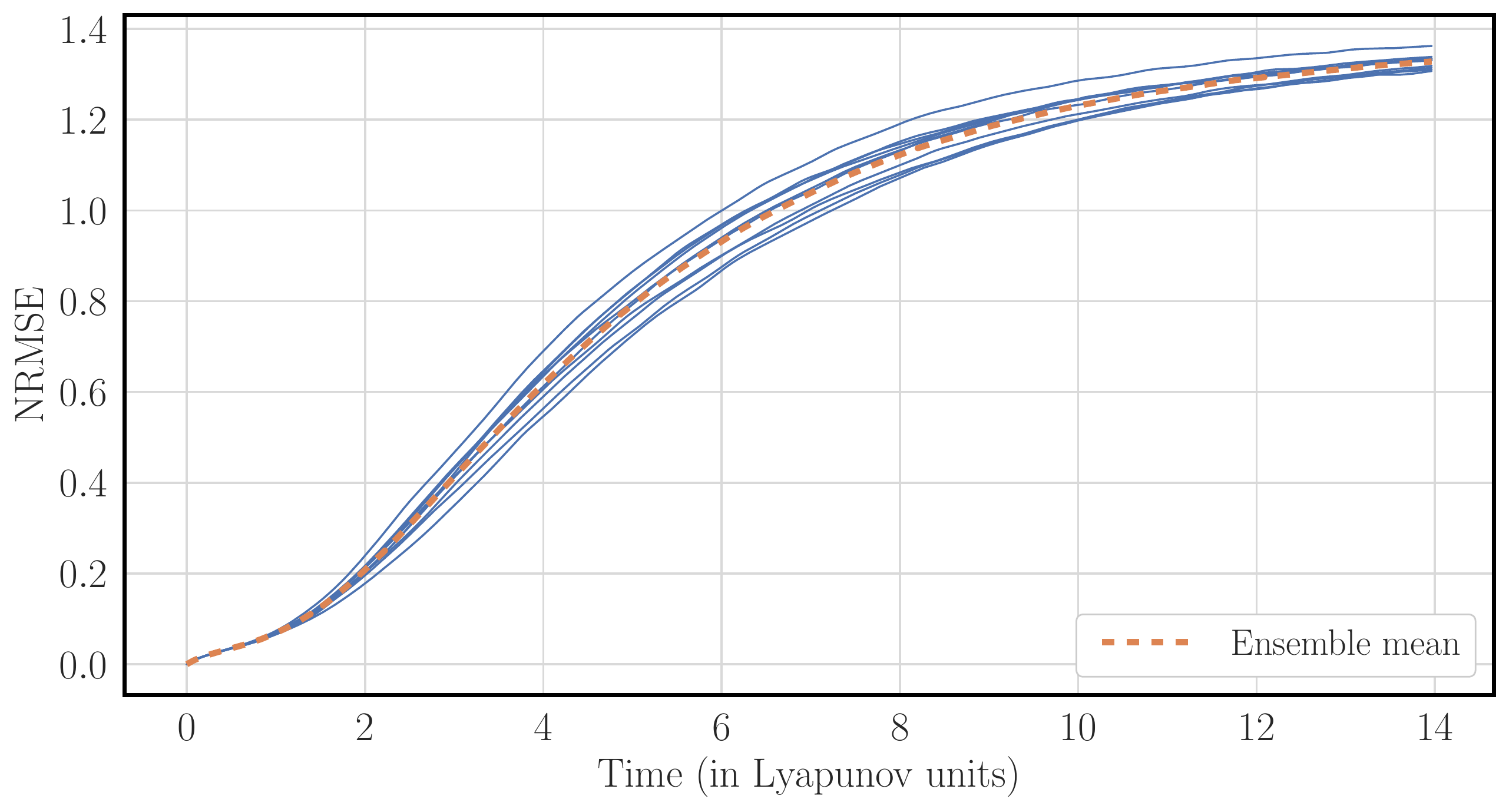} \\
  \includegraphics[width=0.5\textwidth]{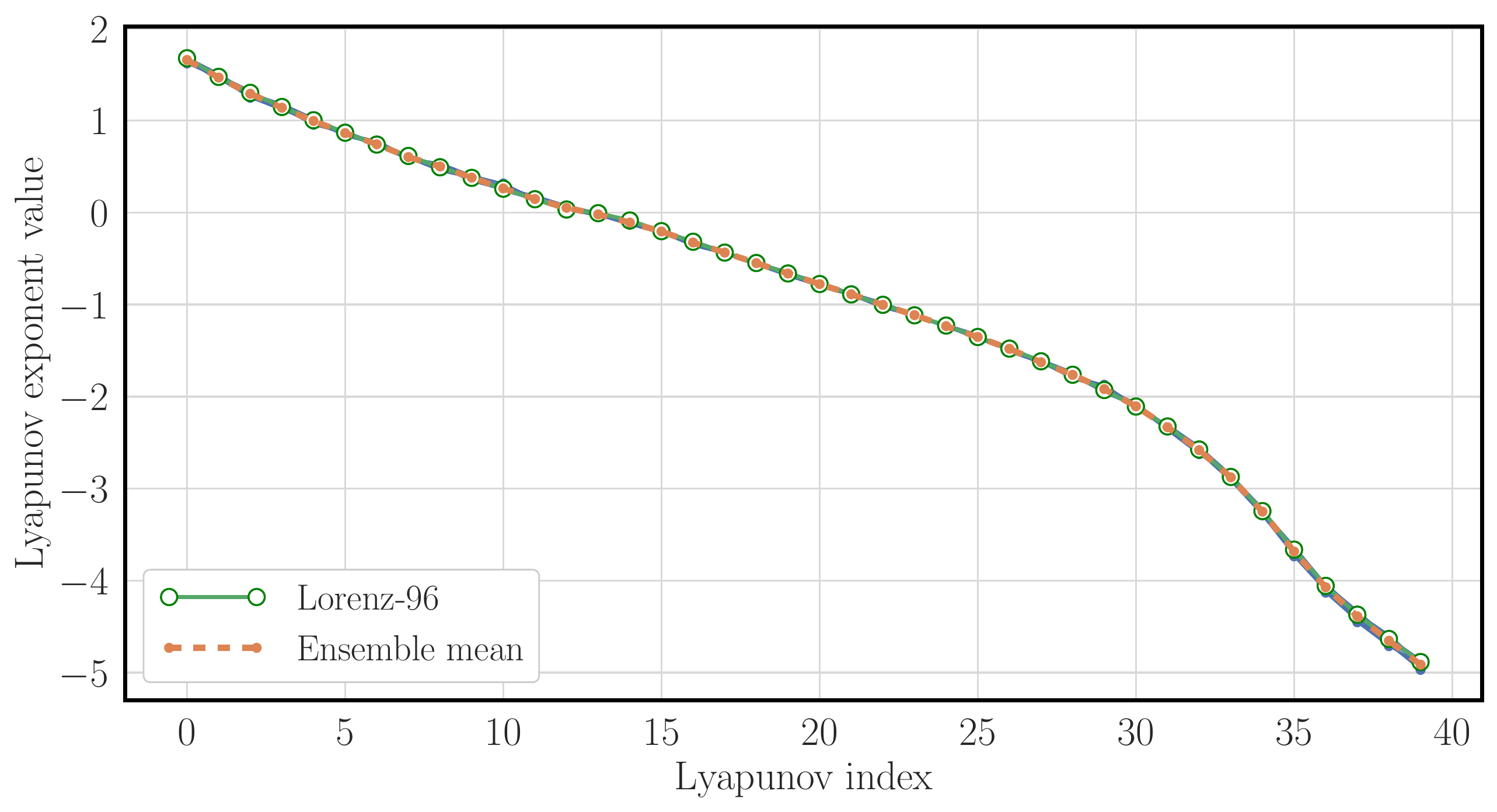} & \includegraphics[width=0.5\textwidth]{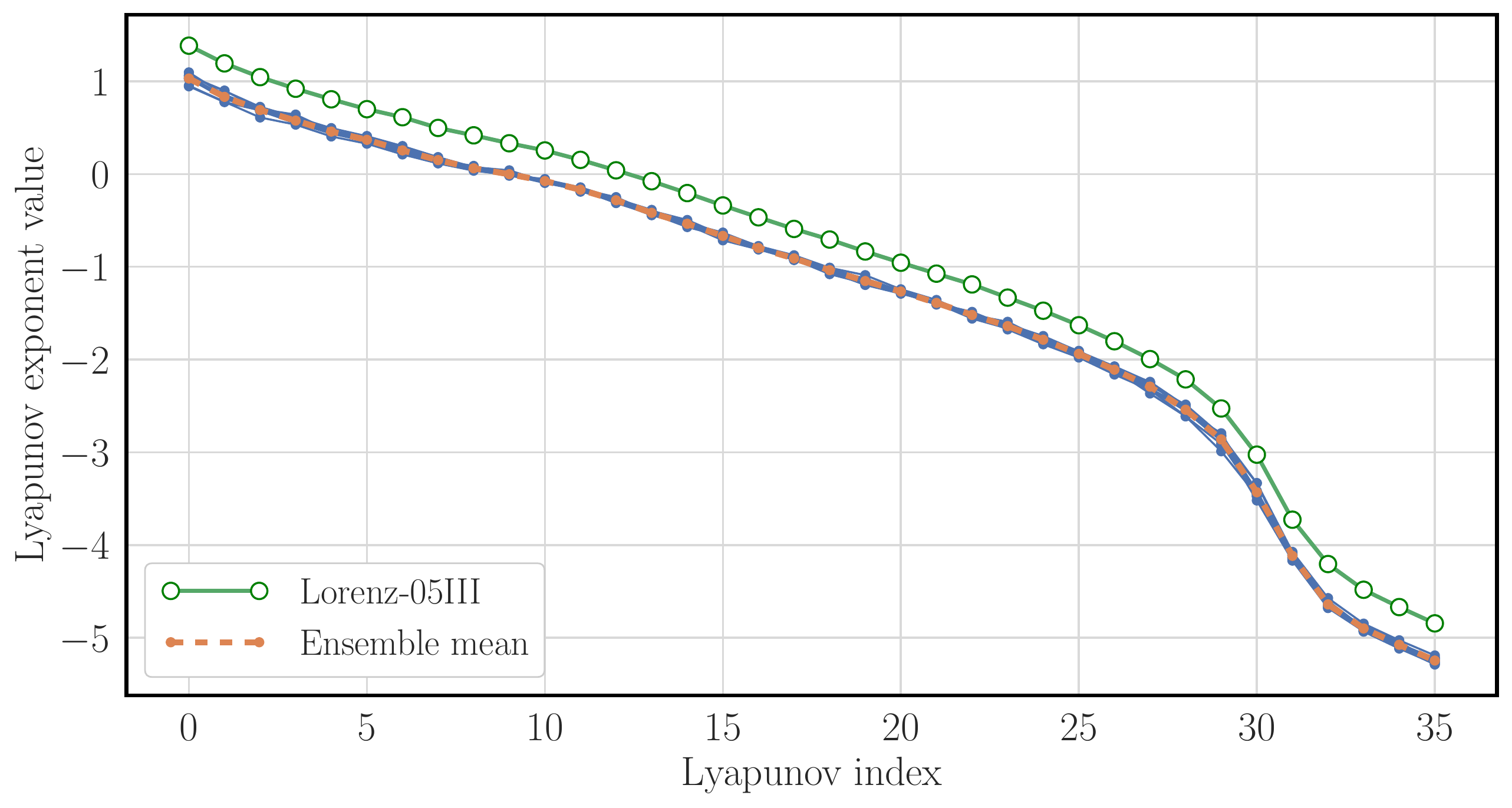} \\
  \includegraphics[width=0.5\textwidth]{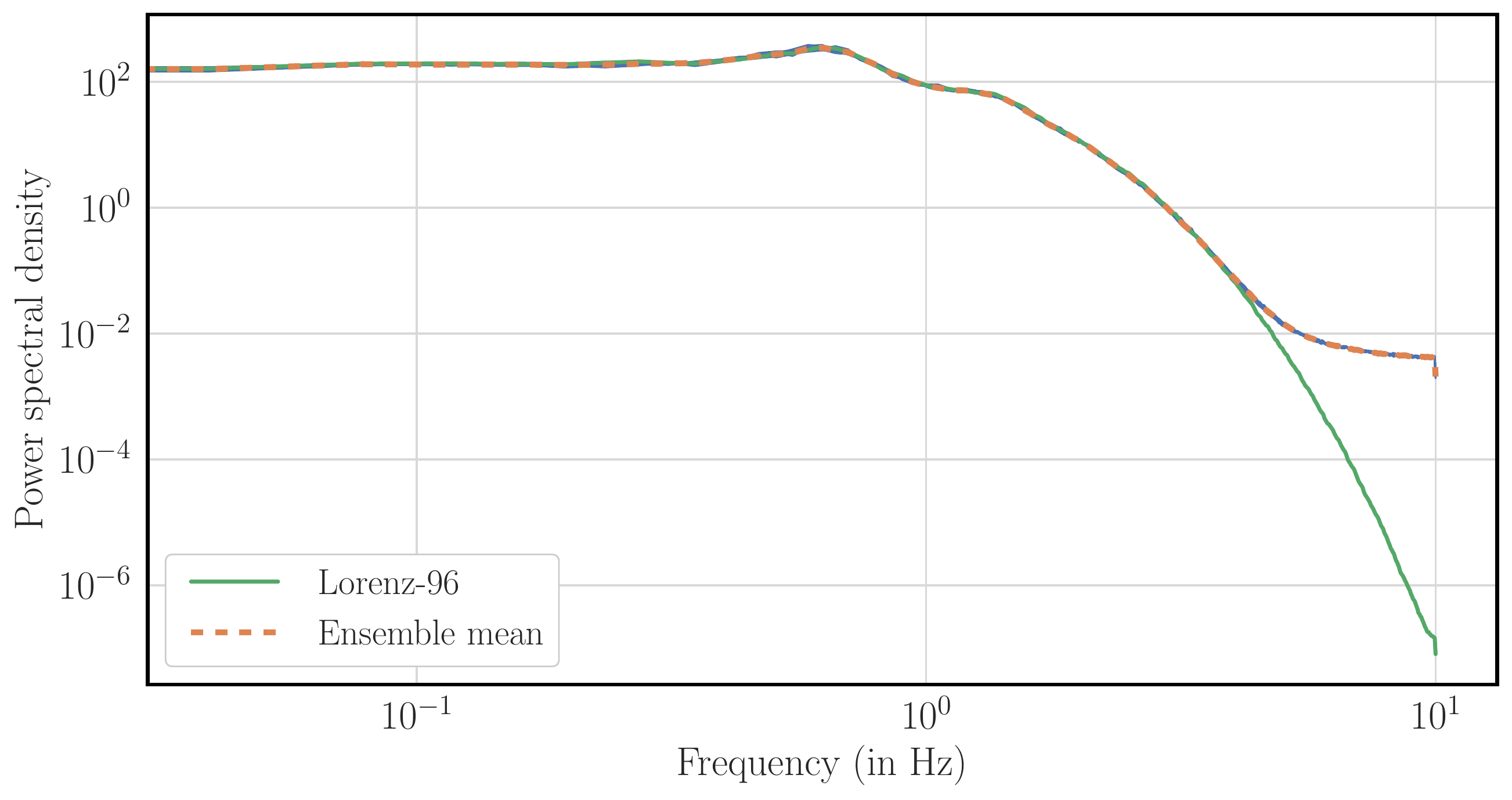} & \includegraphics[width=0.5\textwidth]{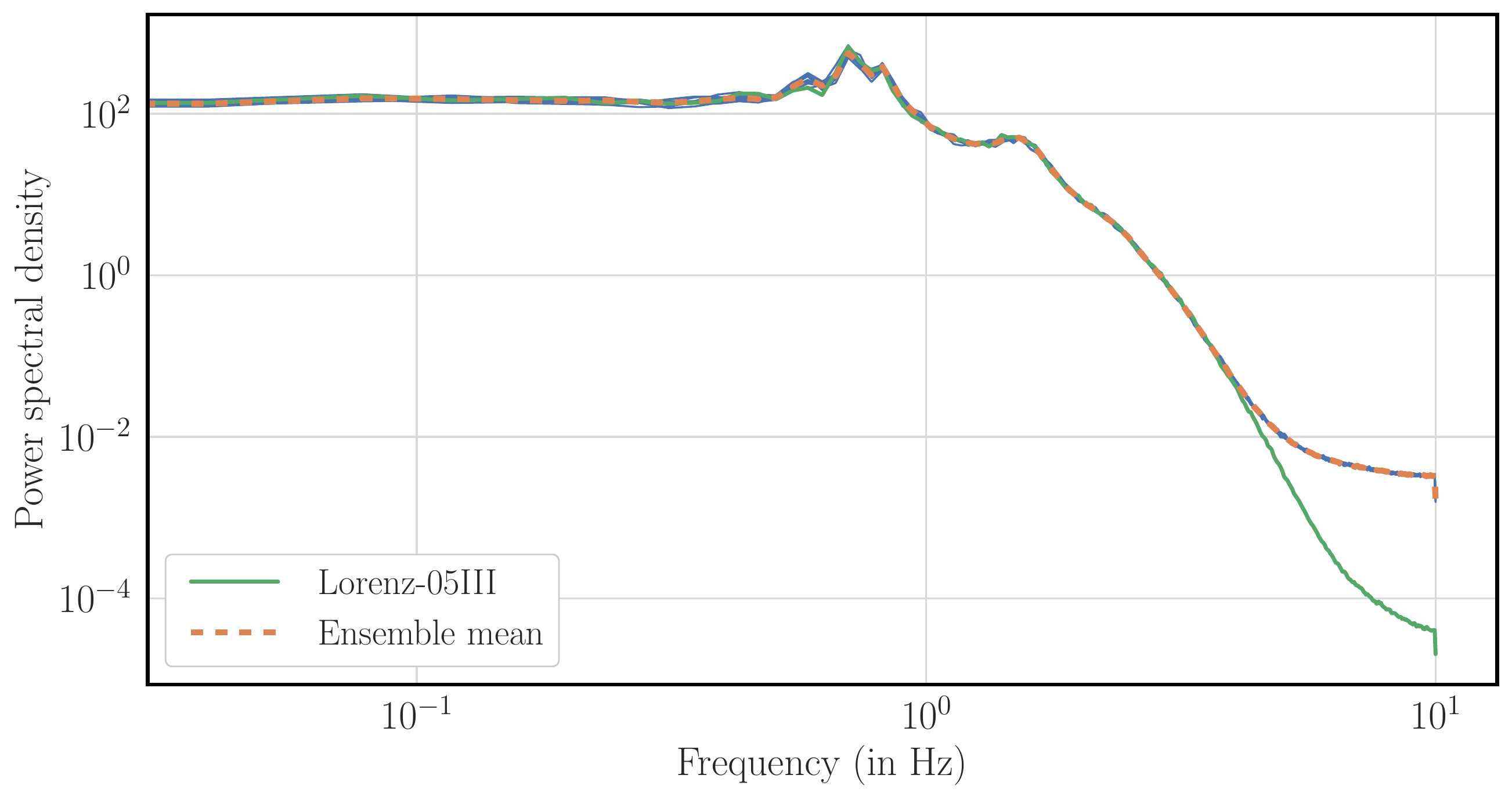}
\end{tabular}
\caption{\label{fig:nominal-l96-l05}
  On the left hand side: Properties of the surrogate model obtained from full but noisy observation of the L96 model in the nominal configuration ($L=4$, $K=5000$, $\sigma_y=1$, $\Ny=\Nx=40$). On the right hand side: Properties of the surrogate model obtained from full but noisy observation of the L05III model in the nominal configuration ($L=4$, $K=5000$, $\sigma_y=1$, $\Ny=\Nx=36$).
  From top to bottom, are plotted the FS (NRMSE as a function of lead time in Lyapunov time), the LS (all exponents), and the PSD (in log-log-scale). A total of $10$ experiments have been performed for both configurations. The curves corresponding to each member are drawn with thin blue lines while the mean of each indicator over the ensemble are drawn in thick dashed orange line.  
}
\end{figure}

The results for the L96 reference model in the nominal configuration with the approximate learning scheme described in Section \ref{sec:approximate-scheme} are reported in Figure~\ref{fig:nominal-l96-l05}, left column, where the FS, the LS, and the PSD are plotted.

We first observe some variability of the FS with the sample, which is small but non negligible.
The asymptotic properties (LS and PSD) are however essentially insensitive to it.
Then LS and PSD of the surrogate model are very close to the reference model ones, which shows that the
surrogate model have very reliable asymptotic properties.
The only discrepancy is seen for the PSD at high frequency where the surrogate model PSD saturates.
This is due to the fact that, for the PSD, we implemented the surrogate model with the additive stochastic noise inferred from the training phase. This quantifies how reliable is the surrogate model depending on the frequency.
Had we only use the deterministic part of the surrogate model, the PSD curves would have actually coincided even better at higher frequency (checked but not shown) but this match at high frequencies would be misleading.

Several scalar indicators are also given in Table \ref{tab:scalar-indicators}.

\begin{table}
  \begin{tabular}{c|cccc|ccc}
    \hline
    Model &  $\Ny$ & $\sigma_y$ & $K$ & $L$ & $\pi_\onehalf$ & $\sigma_q$ & $\lambda_1$  \\
    \hline
    L96 & $40$ & $1$ & $5000$ & $4$ & $4.56 \pm 0.06$ &  $0.08790 \pm 2\, 10^{-5}$ & $1.66 \pm 0.02$ \\
    L05III & $36$ & $1$ & $5000$ & $4$ & $4.06 \pm 0.21$ &  $0.07720 \pm 2\, 10^{-5}$ & $1.03 \pm 0.05$ \\
    \hline
  \end{tabular}
    \caption{\label{tab:scalar-indicators} Scalar indicators for nominal experiments based on L96 and L05III. Key hyperparameters are recalled. The statistics of the indicators are obtained over $10$ samples.}
\end{table}

\subsubsection{Results for the L05III reference model}
The results for the L05III reference model in the nominal configuration with the approximate learning scheme are reported in Figure~\ref{fig:nominal-l96-l05}, right column, where the FS, the LS, and the PSD are plotted.
We observe a higher variability of all three indicators compared to L96.
Even though the sensitivity of LS and FSD remain weak, it is significantly higher for the FS.
The FS shows a steeper curve for short term prediction compared to L96, which is explained by the weaker identifiability of the L05III reference model.
However it relaxes to the climatology at a slower rate compared to L96. This is due to the heterogeneity of the prediction: in the absence of fast mode burst (in space and time) into the slow compartment, the predictability can be long, or it can be shortened whenever the fast variables impact the slow variables. This has been explained and illustrated in \cite{bocquet2019}.

The LS of the surrogate model follows closely that of the reference model but is systematically smaller.
This may be due to the lack of variability of the surrogate model, where only the average impact of the fast variables are learned from the reference model.

The PSD of the surrogate and reference model are close to each other except for the high frequencies.
Similarly to the L96 case, at high frequency, the surrogate model PSD saturates, because the additive stochastic noise inferred from the training phase is accounted for. In the absence of the noise, we checked that there would be a saturation at much higher frequency, but that the PSD of the surrogate model would be offset compared to that of the reference model. This confirms that the stochastic correction plays the important role of stochastic subgrid parameterization in the L05III case.

Several scalar indicators are also given in Table \ref{tab:scalar-indicators}.

\subsection{Comparison of full and approximate scheme}

We have compared the efficiency of the full and approximate schemes for both reference models in their nominal configuration.
First of all, the full scheme is computationally much more demanding: it requires a storage at least $\Ne$ larger than the approximate one and it requires the computation of the cost function \eqref{eq:full-em} which is $\Ne$ times more demanding than cost function \eqref{eq:A-em} of the approximate scheme. In the L96 case, the results for the full and approximate schemes are very close to each other. The FS of the approximate scheme turns out to be slightly better,
while the LS and PSD are almost identical.
Compared scalar indicators are given in Table \ref{tab:full-approx}.

\begin{table}
  \begin{tabular}{cc|ccc}
    \hline
    Model & Scheme & $\pi_\onehalf$ & $\sigma_q$ & $\lambda_1$  \\
    \hline
    L96 &  Approximate & $4.56 \pm 0.06$ &  $0.08790 \pm 2\, 10^{-5}$ & $1.66 \pm 0.02$ \\
    L96 & Full    & $4.24 \pm 0.07$ &  $0.09152$ & $1.66 \pm 0.02$ \\
    \hline
    L05III & Approximate & $4.06 \pm 0.21$ &  $0.07720 \pm 2\, 10^{-5}$ & $1.03 \pm 0.05$ \\
    L05III & Full & $3.97 \pm 0.17$ &  $0.08024 $ & $1.03 \pm 0.04$ \\
    \hline
  \end{tabular}
    \caption{\label{tab:full-approx} Scalar indicators for L96 and L05III in their nominal configuration, using either the full or the approximate schemes. The statistics of the indicators are obtained over $10$ samples.}
\end{table}

In the L05III case, the LS and PSD curves are also very close to each other.
However, the FS of the approximate scheme is distinctively better than the FS of the full scheme as summarized by $\pi_\onehalf$ in Table \ref{tab:full-approx}.

The edge that the approximate scheme has on the full scheme as far as FS is concerned can be explained as follows.
The approximate scheme looks for $\bA$ as the MAP rather than the average as the full scheme would.
Hence, it has the tendency to improve the short term deterministic FS as much as possible and hence minimizes $\sigma_q$, at the risk of overfitting to the FS criterion. But it might be less reliable in terms of the representation of the uncertainty.
This effect is more noticeable in the L96 case. It could be due to the better identifiability of the L96 reference model by the surrogate model.

\subsection{Sensitivity on the length of the training window}

One key advantage of the schemes proposed in Section \ref{sec:optimization-em} resides in their ability to handle long training windows.
Note that a clear tendency when increasing the length $K$ of the training windows was difficult to exhibit in \cite{bocquet2019,brajard2020}. 
Here, we consider the nominal configurations for L96 and L05III, i.e. with $L=4$, but with
\begin{equation}
K \in \left\{50, 100, 200, 400, 800, 1600, 3200, 6400\right\} .
\end{equation}
Moreover, for each configuration,
average indicators are obtained from $10$ sample runs (each one with a different random seed).

\begin{figure}
\begin{tabular}{ccc}
  \includegraphics[width=0.5\textwidth]{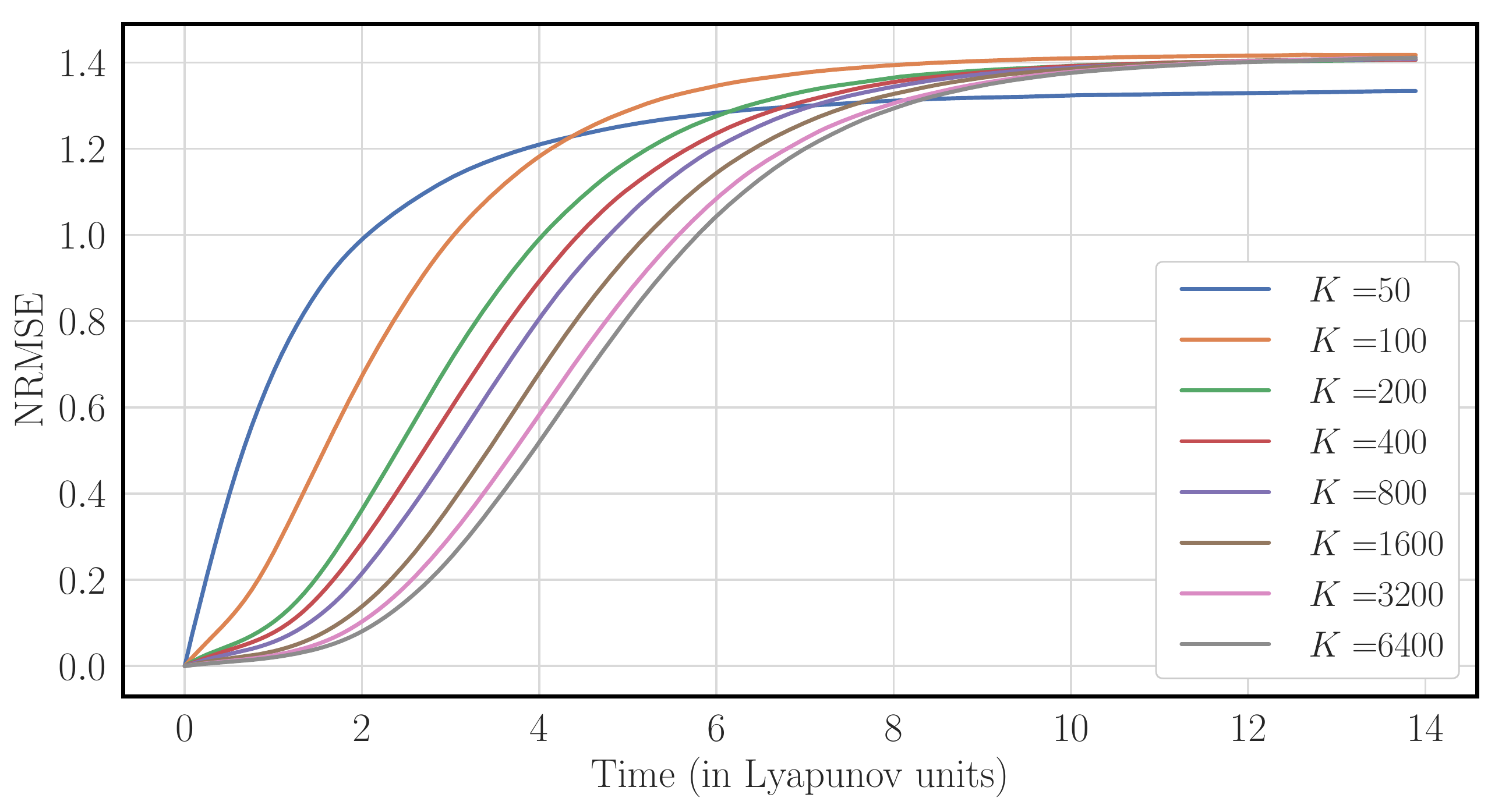} & \includegraphics[width=0.5\textwidth]{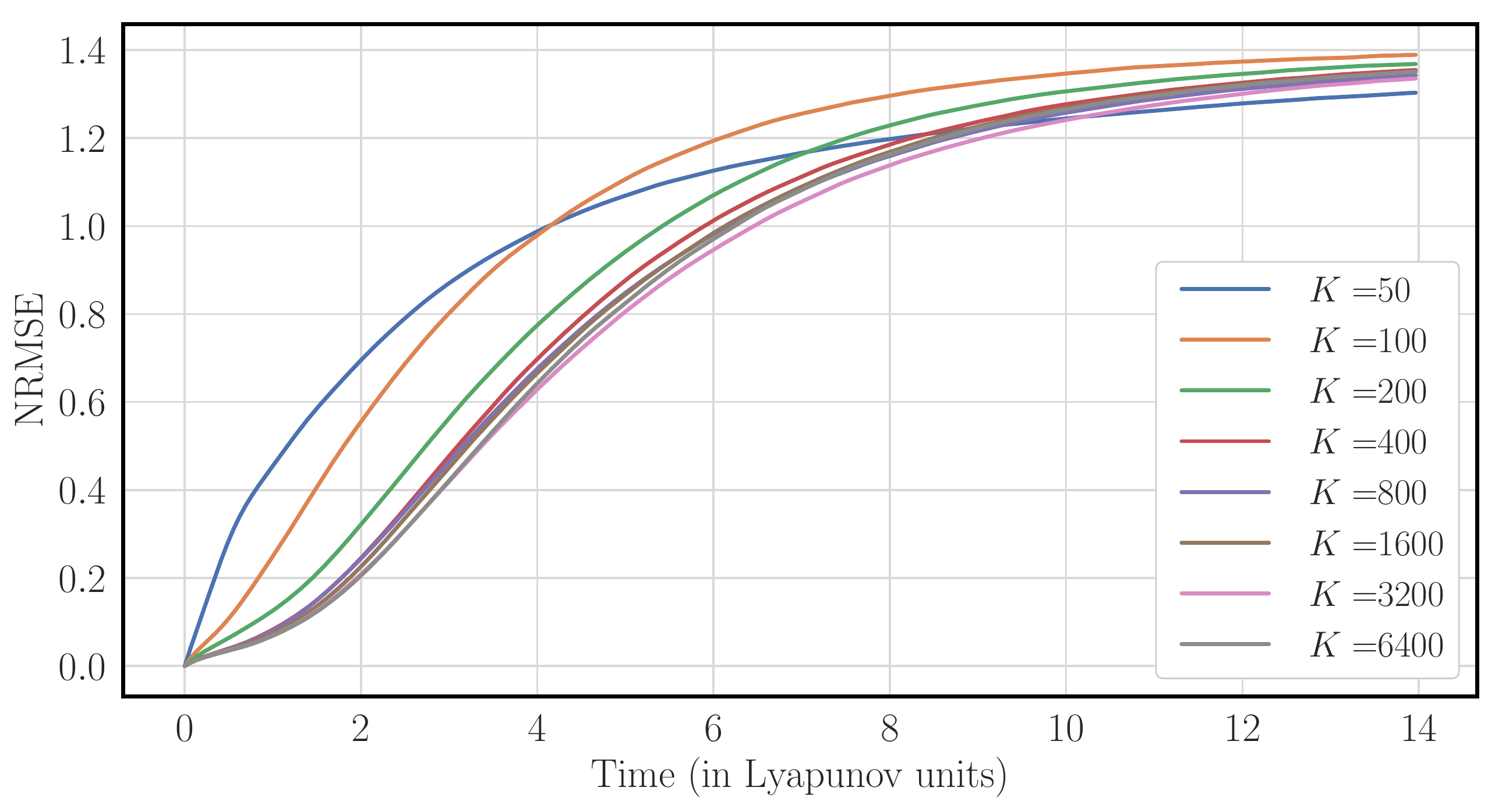} \\
  \includegraphics[width=0.5\textwidth]{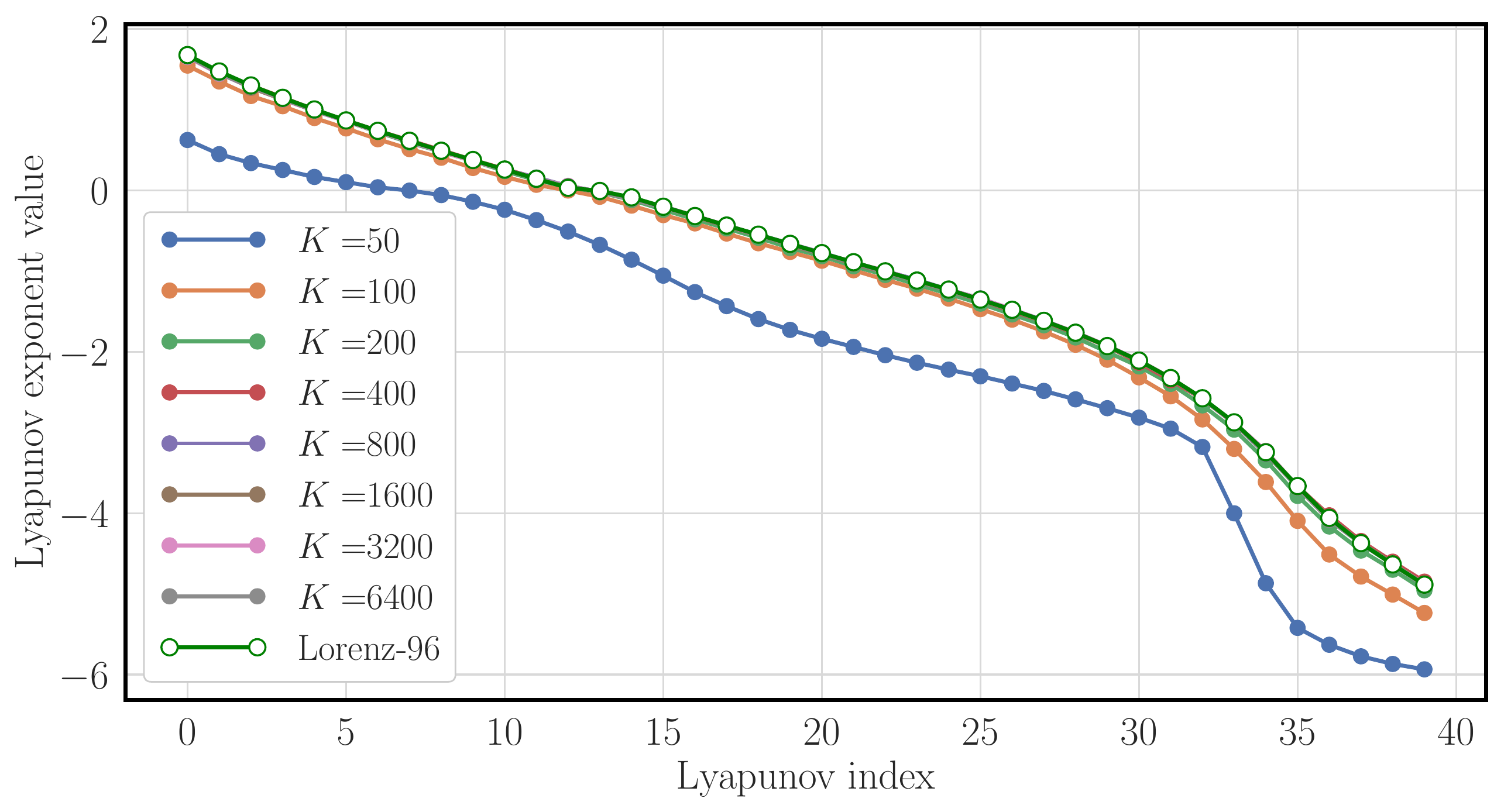} & \includegraphics[width=0.5\textwidth]{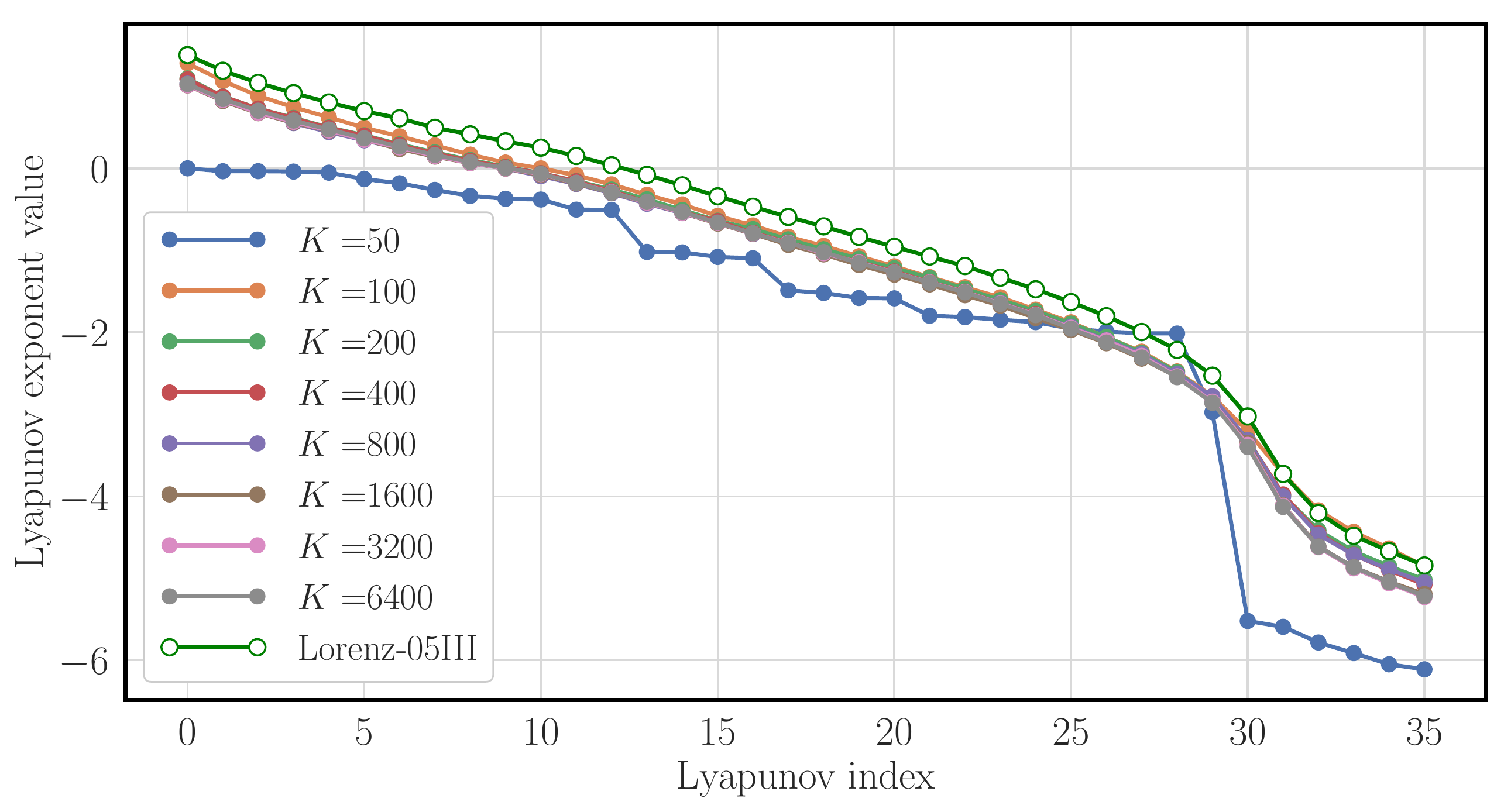} \\
  \includegraphics[width=0.5\textwidth]{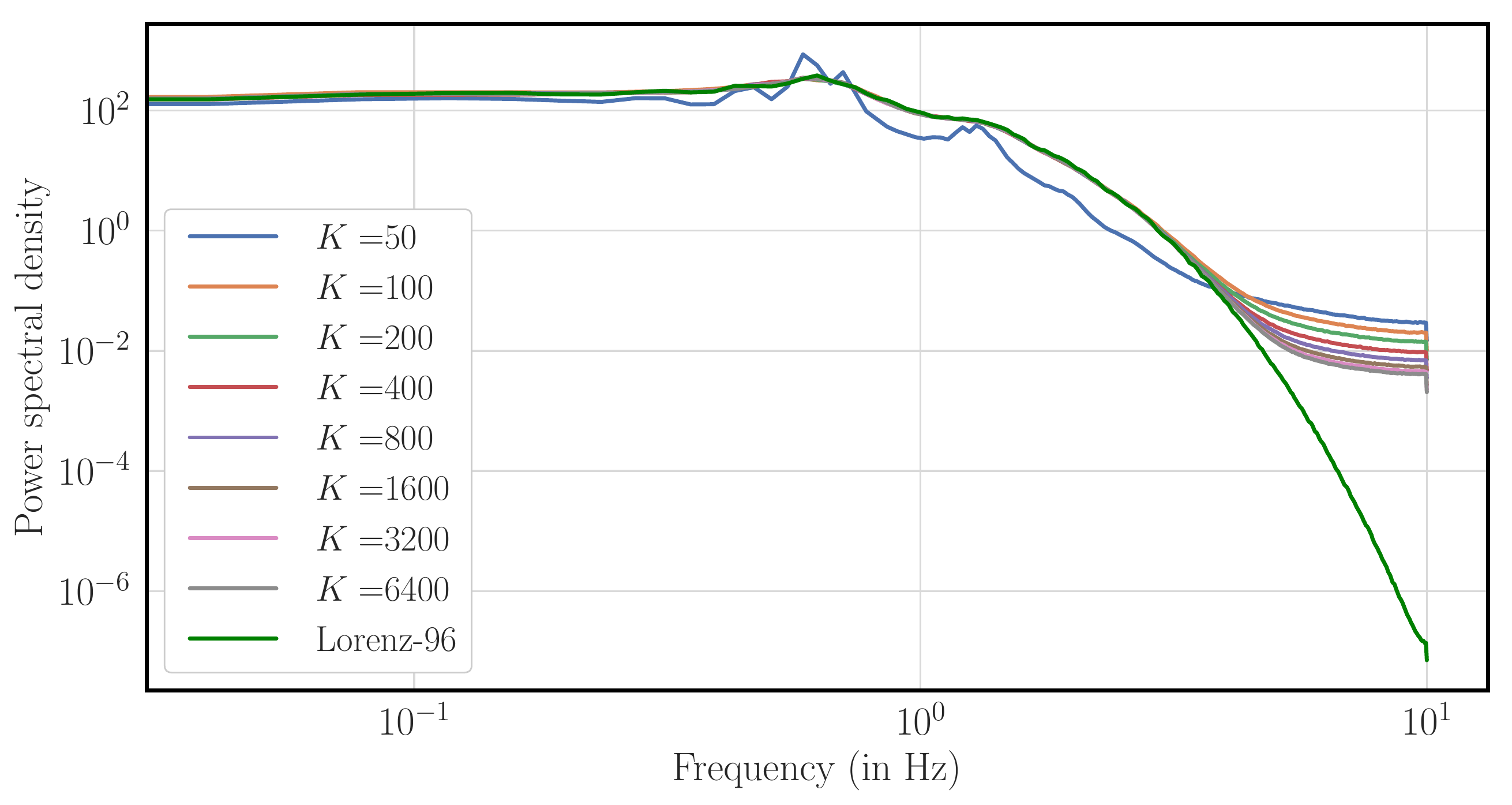} & \includegraphics[width=0.5\textwidth]{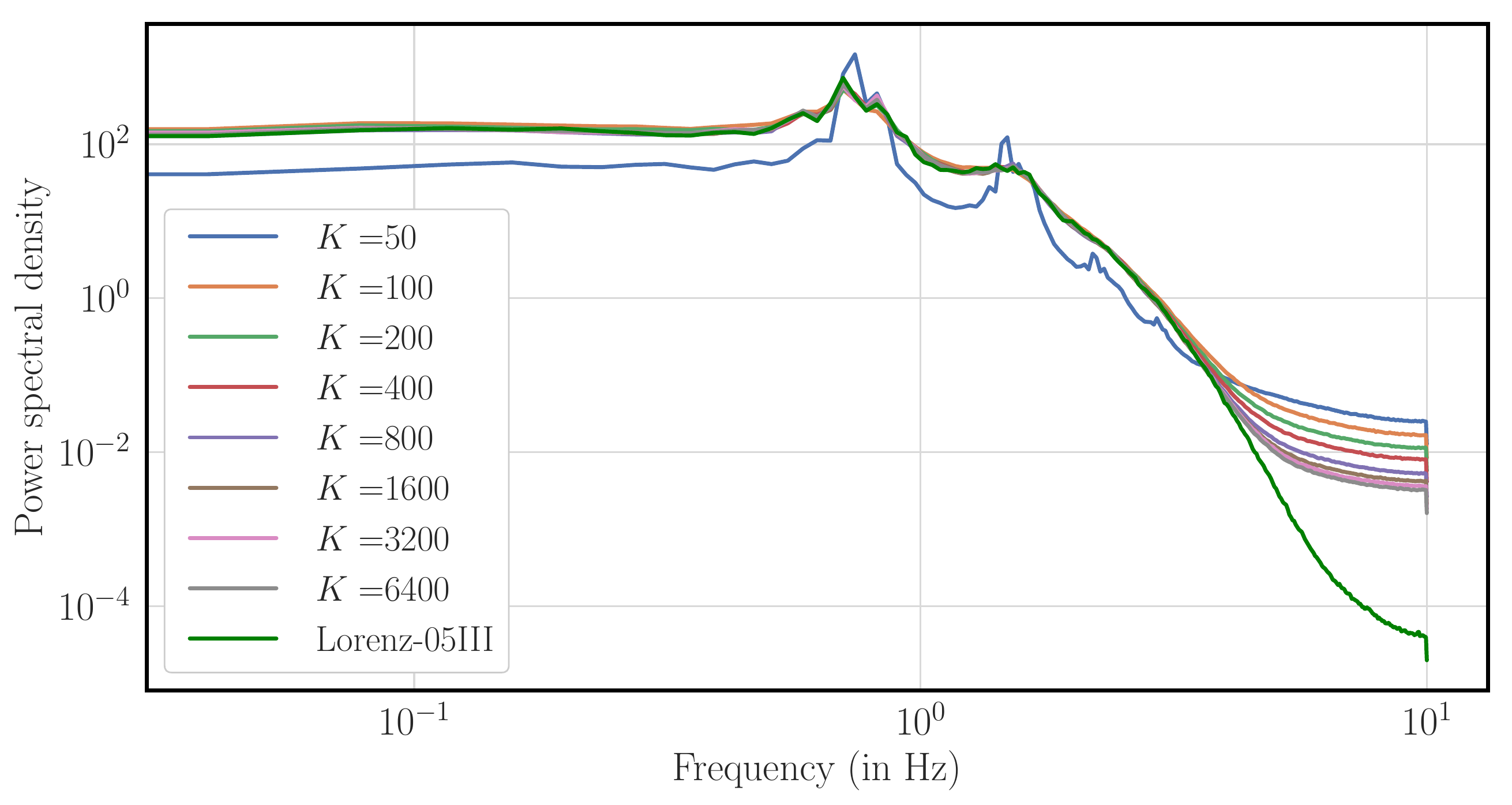}
\end{tabular}
\caption{\label{fig:K-dependence}
  Same as Figure~\ref{fig:nominal-l96-l05} but for several values of the training window length $K$.
  Each curve is the mean over $10$ experiments with different sets of observations. 
  The LS and PSD of the reference models are also plotted for comparison.
}
\end{figure}

The results are plotted in Figure~\ref{fig:K-dependence}, following the same structure as Figure~\ref{fig:nominal-l96-l05},
except that we now only plot the means but for several $K$ values.
It can be checked in the L96 case that the FS improves for all lead times with increasing $K$, as expected.
The tendency is also verified in the L05III case, but not as clearly for large $K$. It may be due to the higher variability of the FS in the L05III case observed previously. But this is more likely due to a saturation of the performance as $K$ increases, due to the weaker identifiability of the L05III model compared to that of the L96 model.
For both models, the surrogate model LS has converged with good accuracy whenever $K \ge 400$,
and the PSD converges to that of the reference model with good accuracy whenever $K \ge 200$, except for the high frequencies. Let us remark that the high frequency PSD is recovered better and better as $K$ increases.

\subsection{Sensitivity on the lag of the smoother}

We have carried out experiments keeping $K=5000$, but with the lag of the smoother chosen in
\begin{equation}
L \in \left\{ 0, 2, 4, 6, 8, 10, 12 \right\}.
\end{equation}
For each configuration, $10$ independent samples are obtained.
In the case of L05III, $L=4$ yields the best FS, while $L=2,4$ are optimal for the FS in the case of L96.
Note that in both cases, the configuration $L=0$, i.e. filtering in place of smoothing, significantly degrades the FS.
The LS and PSD are improved with $L\ge 2$ but only marginally.
The reasons for these results had been tentatively given in Section \ref{sec:prelim}.

\subsection{Sensitivity on the observation noise}

\begin{figure}
\begin{tabular}{ccc}
  \includegraphics[width=0.5\textwidth]{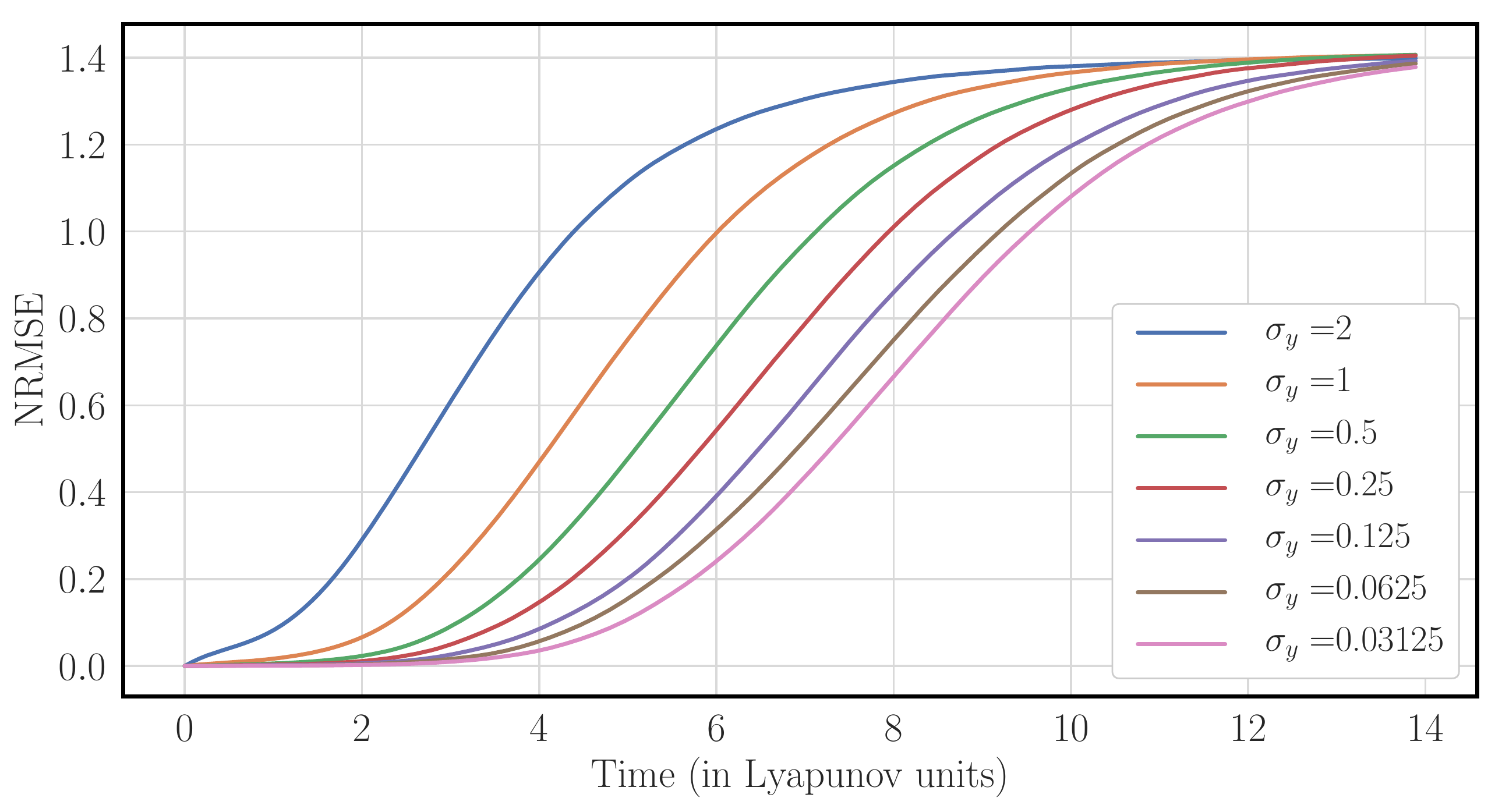} & \includegraphics[width=0.5\textwidth]{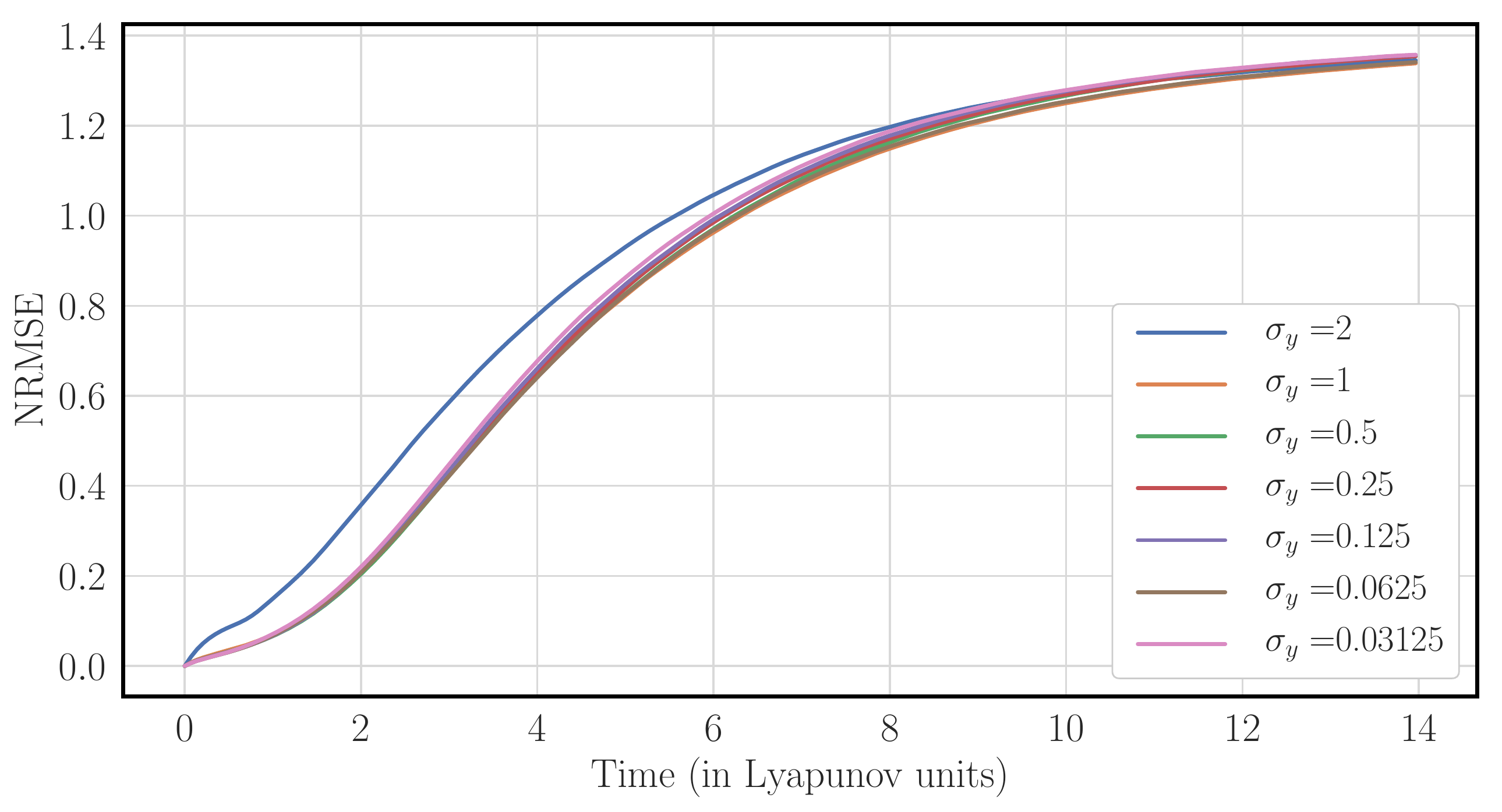} \\
  \includegraphics[width=0.5\textwidth]{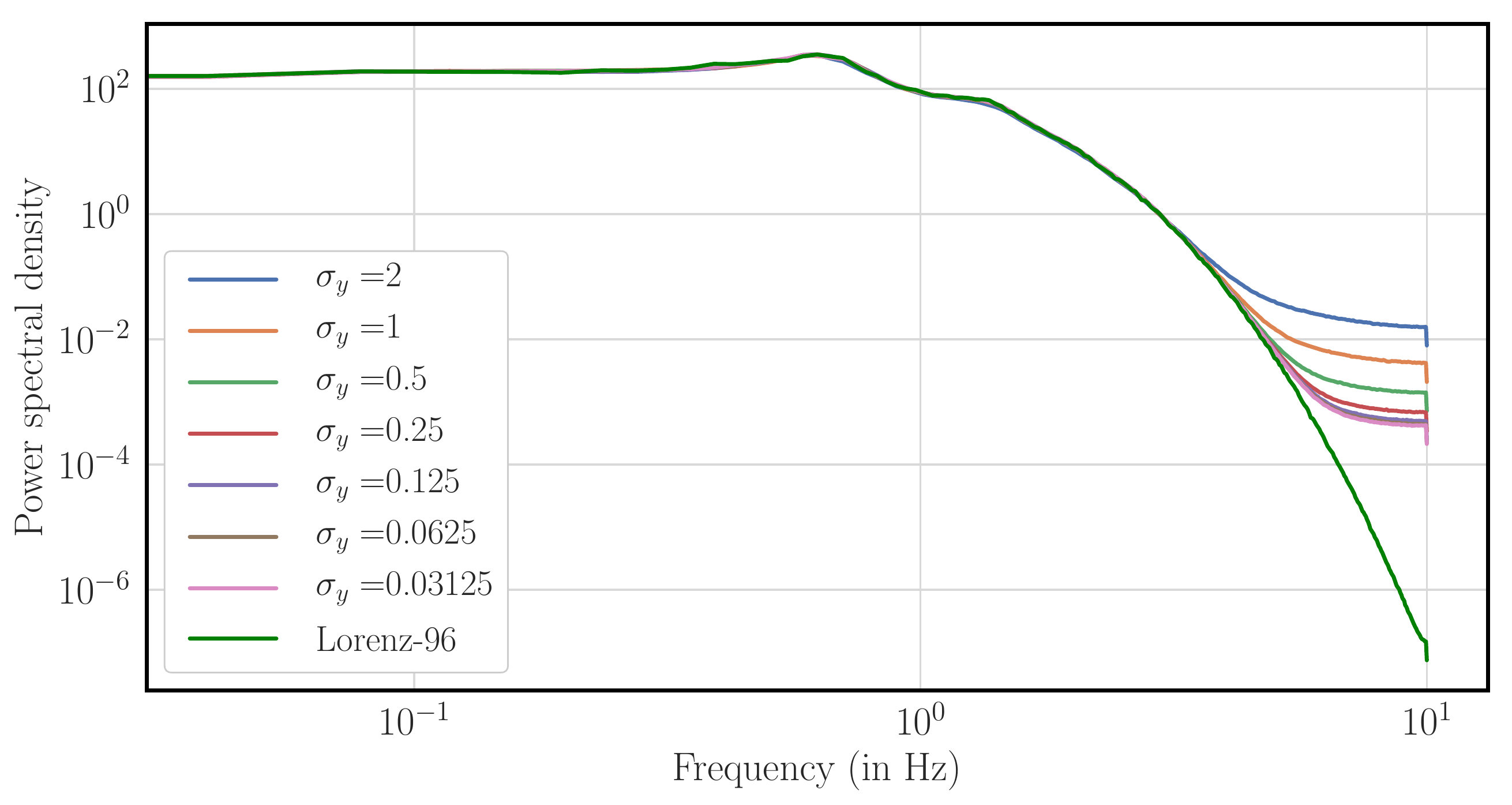} & \includegraphics[width=0.5\textwidth]{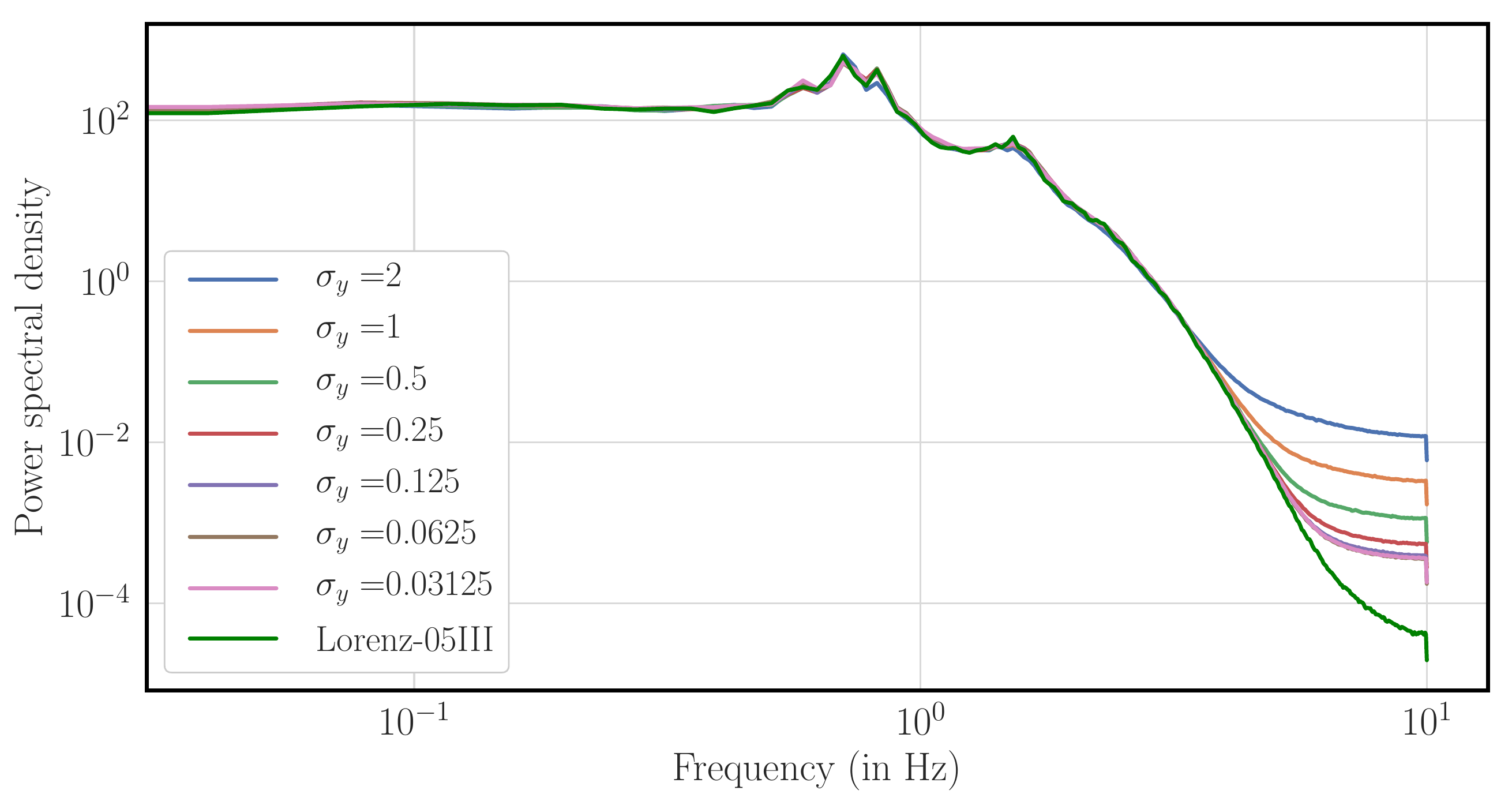}
\end{tabular}
\caption{\label{fig:sigma_y}
  On the left hand side: Properties of the surrogate model obtained from full but noisy observation of the L96 model in the nominal configuration ($L=4$, $K=5000$, $\Ny=\Nx=40$ and with several $\sigma_y$). On the right hand side: Properties of the surrogate model obtained from full but noisy observation of the L05III model in the nominal configuration ($L=4$, $K=5000$, $\sigma_y=1$, $\Ny=\Nx=36$ and with several $\sigma_y$).
  From top to bottom, are plotted the FS (NRMSE as a function of lead time in Lyapunov time) and the PSD (in log-log-scale), averaged over an ensemble of $10$ samples.
}
\end{figure}

We also investigate the dependence on the observation noise magnitude.
For both models, in the nominal configurations, the observation error standard deviation (recall $\bR = \sigma_y^2 \Ix$) takes value in
\begin{equation}
\sigma_y = \left\{2, 1 , 2^{-1}, 2^{-2}, 2^{-3}, 2^{-4}, 2^{-5}\right\} .
\end{equation}
The results for the FS and PSD are reported in Figure \ref{fig:sigma_y}.
In the L96 case, the FS keeps increasing as $\sigma_y$ decreases.
This confirms the good identifiability of the L96 model by the surrogate model's architecture.
In contrast, in the L05III case, the FS saturates with decreasing observation noise magnitude as soon as $\sigma_y=1$, which is due to the mild identifiability given the chosen structure for the surrogate model.
For both reference models, however, the PSD keeps improving at high frequencies as $\sigma_y$ decreases.

\subsection{Sensitivity on the density of the observations}

\begin{figure}
\begin{tabular}{ccc}
  \includegraphics[width=0.5\textwidth]{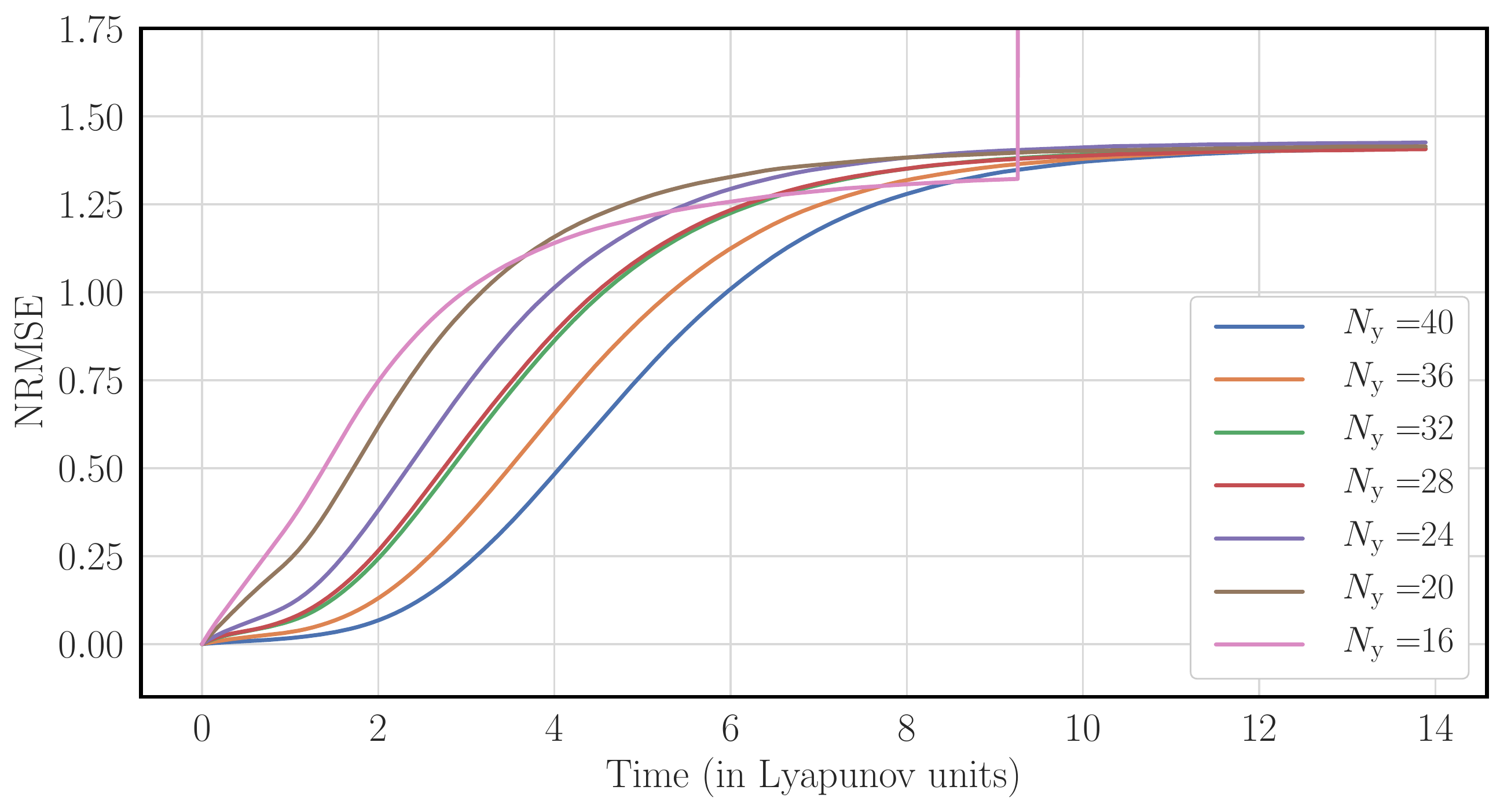} & \includegraphics[width=0.5\textwidth]{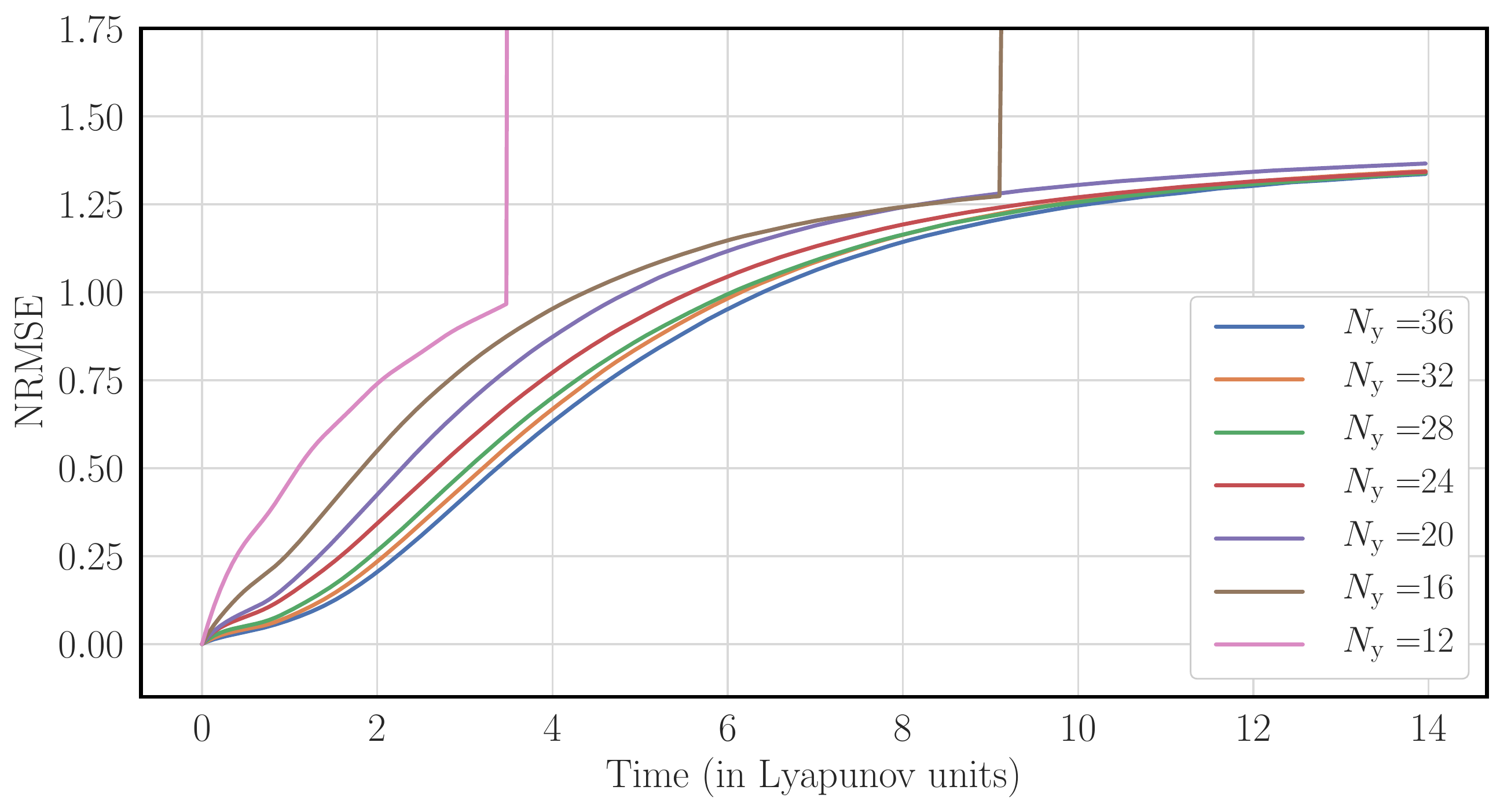} \\
  \includegraphics[width=0.5\textwidth]{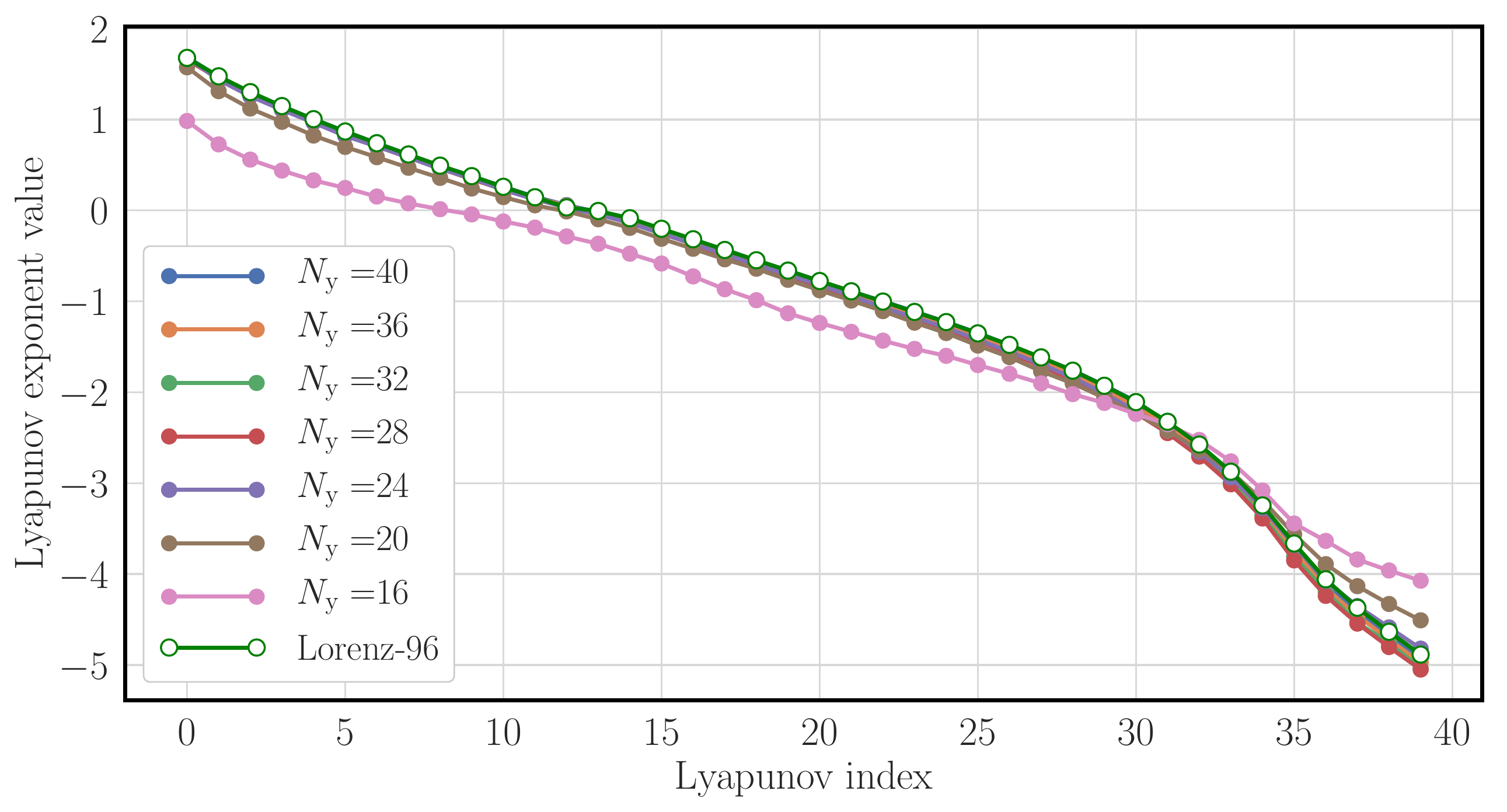} & \includegraphics[width=0.5\textwidth]{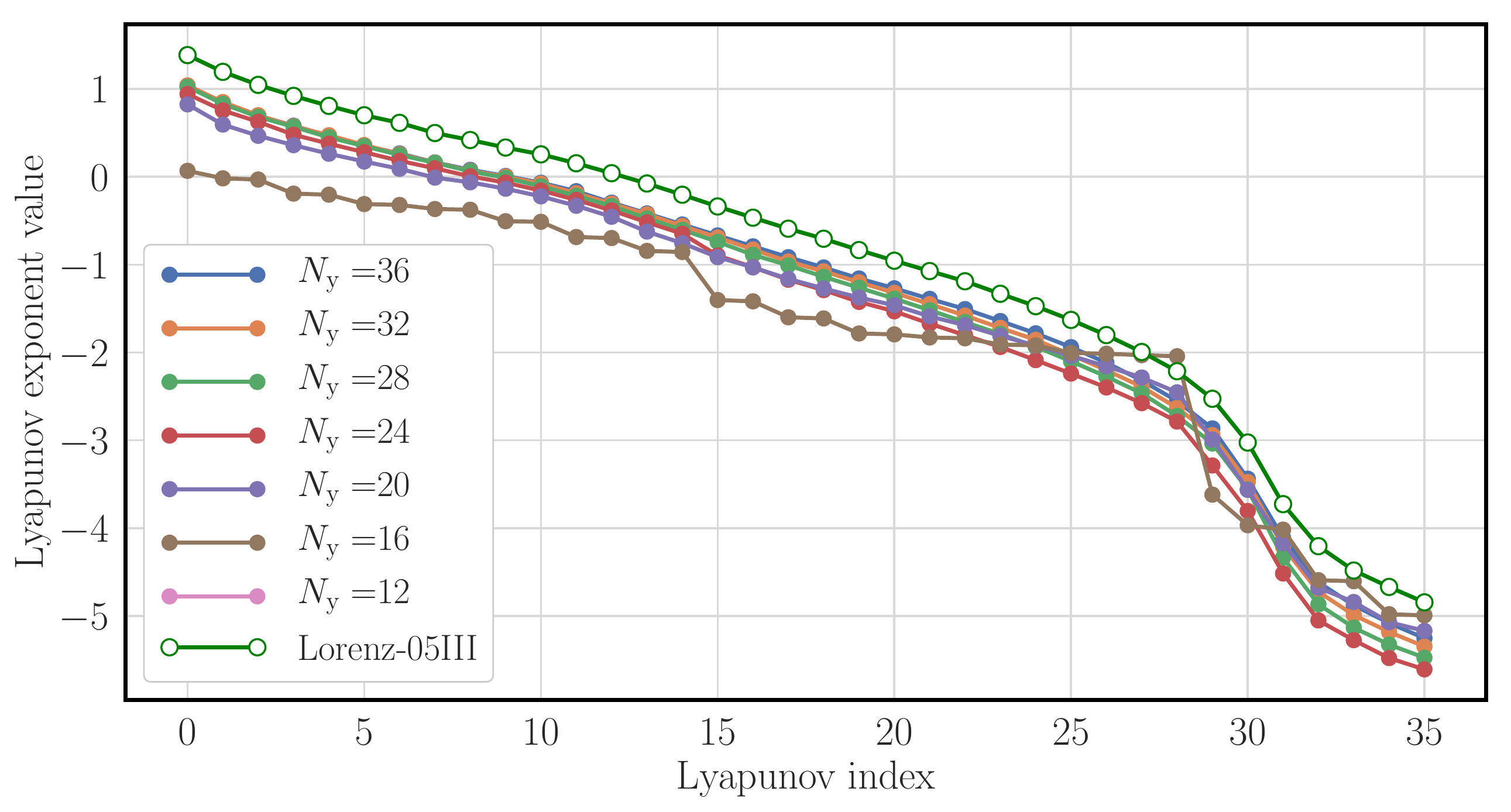} \\
  \includegraphics[width=0.5\textwidth]{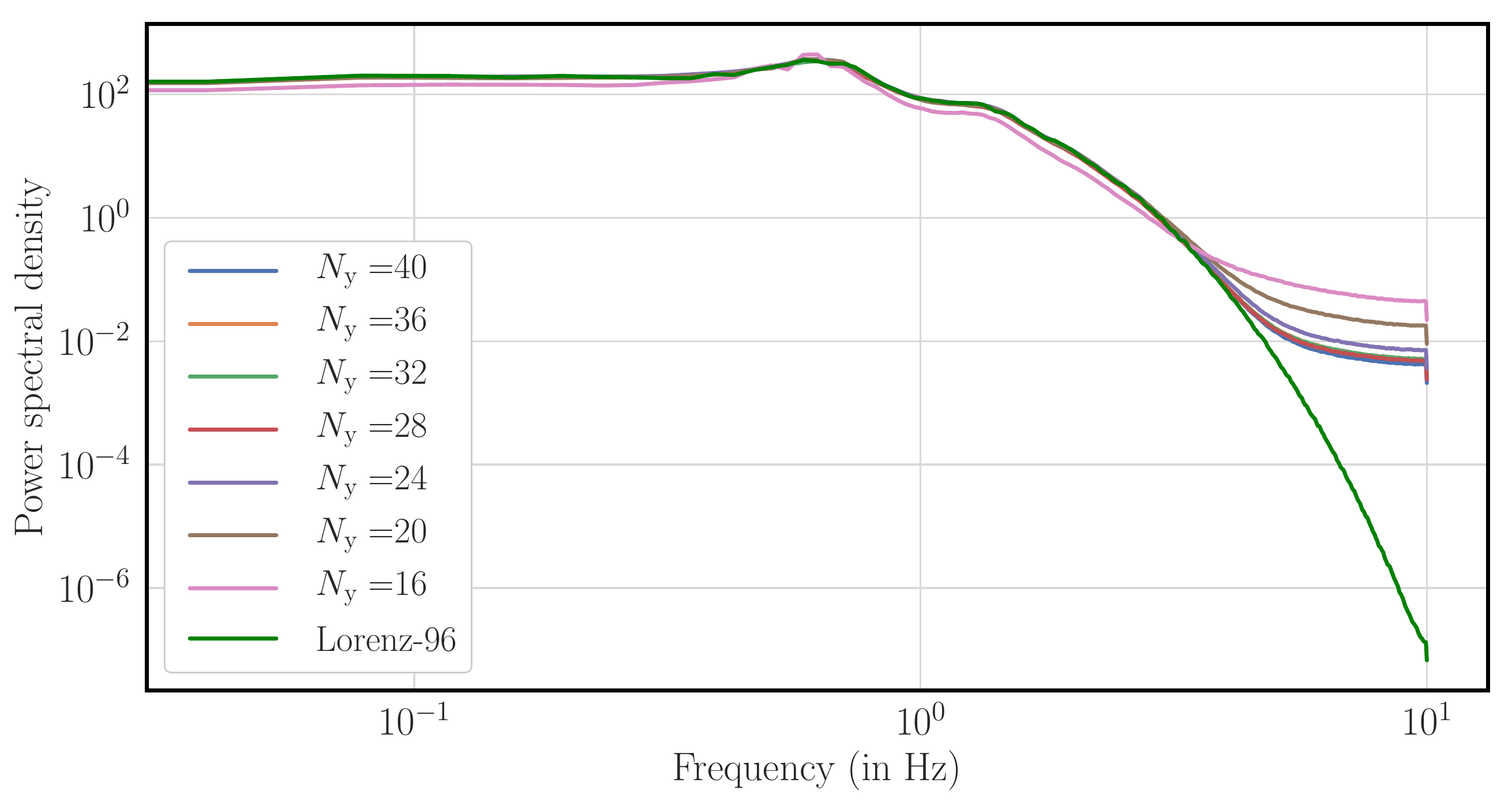} & \includegraphics[width=0.5\textwidth]{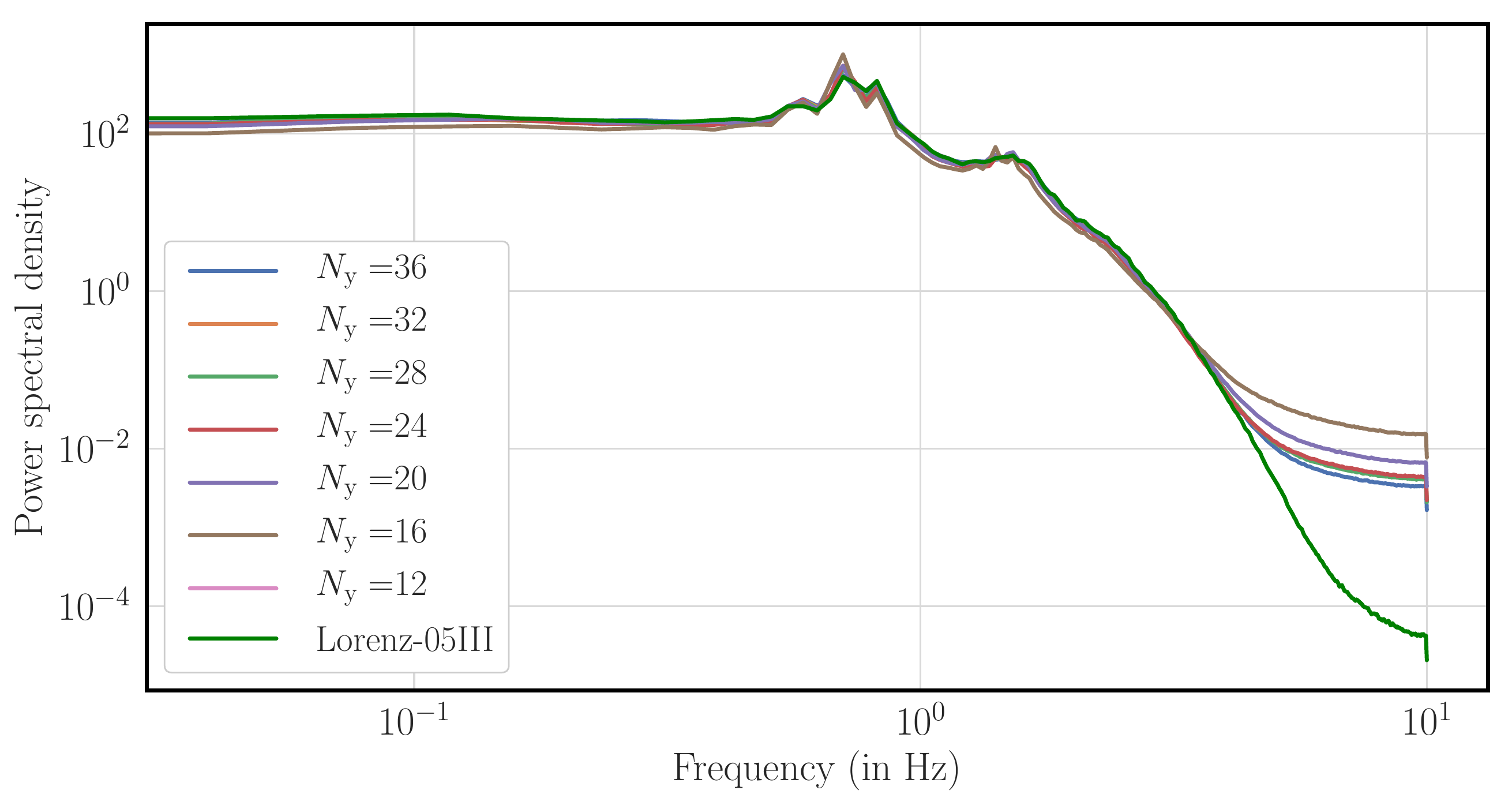}
\end{tabular}
\caption{\label{fig:Ny}
  On the left hand side: Properties of the surrogate model obtained from partial and noisy observation of the L96 model in the nominal configuration ($L=4$, $K=5000$, $\sigma_y=1$, $\Nx=40$) where $\Ny$ is varied. On the right hand side: Properties of the surrogate model obtained from partial and noisy observation of the L05III model in the nominal configuration ($L=4$, $K=5000$, $\sigma_y=1$, $\Nx=36$) where $\Ny$ is varied.
  From top to bottom, are plotted the mean FS (NRMSE as a function of lead time in Lyapunov time), the mean LS (all exponents), and the mean PSD (in log-log-scale). A total of $10$ experiments have been performed for both configurations.
}
\end{figure}

Finally the dependence of the quality of the surrogate model as a function of the density of the monitoring network is investigated. For both models, we consider the nominal configuration but where the monitoring network at $t_k$, consists in the $k$-dependent random uniform draw of $\Ny$ sites where the variables are observed with an unbiased Gaussian noise of standard deviation $\sigma_y=1$.
The number of observations at each observation time step is taken in the set
$
\Ny \in \left\{ 36, 32, 28, 24, 20, 16, 12 \right\},
$
for L05III and in the set
$
\Ny \in \left\{ 40, 36, 32, 28, 24, 20, 16 \right\},
$
for L96. The results for the FS, LS and PSD are reported in Figure \ref{fig:Ny}.
As expected, the performance degrades significantly with the increase of sparsity.
When $\Ny$ is smaller than $\Nx/2$, some of the training steps may fail.
This is most likely due to the inability of the EnKF/EnKS to be stable with higher sparsity, as is well known (see e.g., Section 4.3 in \cite{bocquet2019}).
Localization is known to improve the robustness of the filter/smoother with higher sparsity but to a point.

As in the previous experiments, there is a saturation of the L05III FS performance for dense networks, as opposed to the L96 case which shows a significant improvement even from $\Ny=36$ to $\Ny=40$.
Note that, in contrast with the previous results, the quality of LS and PSD are significantly dependent on the density of the monitoring network, especially in the L05III case.

\section{Conclusion}
 
In this paper, we have proposed a unifying Bayesian view on the extended data assimilation problem where not only the state trajectory is a control variable, but the dynamical model and its model error statistics are as well. This enables to learn about these control variables from partial and noisy observations of the physical system.
The surrogate dynamical model is formalized as a neural network built with convolutional layers leveraging locality and possibly homogeneity assumptions.
Techniques proposed so far to optimize the cost function associated to this problem have been re-interpreted as coordinate descent schemes.
The EM technique was used to solve the marginal problem where the most likely model and model error statistics are both solved for.
The algorithm alternates a traditional data assimilation step based on an EnKS and where a posterior ensemble is obtained, with an optimization step where new iterates of the model and the model error statistics are derived using machine learning tools, a quasi-Newton optimizer and analytical formula. The scheme enables to successfully handle very long training windows.
Note that the outcome is not only a deterministic surrogate model but also its associated stochastic correction, representative of the uncertainty attached to the deterministic part, which accounts for residual model errors.

An approximate scheme where model error statistics are computed online from the ensemble during the EnKS step is proposed and compared to the full EM scheme which requires the ensemble to be stored.
These two schemes are successfully tested on two low-order chaotic models with distinct identifiability, investigating the sensitivity to the length of the training window, the observation error magnitude, the density of the monitoring network, the lag of the EnKS, and using statistical indicators that probe short-term and asymptotic properties of the surrogate model.

This study has answered several of the questions originally raised when these combined data assimilation and machine learning methods were introduced in \cite{bocquet2019,brajard2020}.
A few others remain. How do we incorporate a priori information on the dynamical model?  How do we build the corresponding online learning system as in sequential data assimilation? There are also technical challenges to address, such as those arising when applying these techniques to more complex, higher-dimensional models.

\section*{Acknowledgments}
The authors are grateful to two reviewers for their comments and suggestions that helped improve the paper.
The authors are thankful to Alban Farchi and Quentin Malartic for their suggestions on the original manuscript.
This work was granted access to the HPC resources of IDRIS under the allocation AD011011184 made by GENCI.
AC, JB and LB have been funded by the project REDDA (\#250711) of the Norwegian Research Council.
AC was also supported by the Natural Environment Research Council (Agreement PR140015 between NERC and the National Centre for Earth Observation).
CEREA and LOCEAN are members of Institut Pierre--Simon Laplace (IPSL).



\bibliographystyle{AIMS}
\bibliography{references.bib}

\medskip
Received xxxx 20xx; revised xxxx 20xx.
\medskip

\end{document}